\documentclass[11pt,a4paper]{article}
\usepackage{bbm}
\usepackage{amssymb}
\usepackage{amsmath}
\usepackage{graphicx}
\usepackage{subfigure}
\usepackage{cite}
\usepackage{algorithm}
\usepackage[scale={0.72,0.76}, hmarginratio=1:1, vmarginratio=1:1, headheight=2ex, headsep = 2ex, footskip = 3.6ex, nomarginpar]{geometry}
\usepackage{algorithmic}
\usepackage{color}
\newtheorem{theorem}{Theorem}

\newtheorem{lemma}{Lemma}

\newtheorem{proposition}{Proposition}

\newtheorem{definition}{Definition}
\newtheorem{assumption}{Assumption}

\newcommand{\red}{\color{red}}

\begin{document}
\title{ Deep Neural Networks for Rotation-Invariance Approximation and Learning\thanks{The research of CKC was partially supported by Hong
Kong Research Council [Grant Nos. 12300917 and 12303218] and Hong
Kong Baptist University [Grant No. HKBU-RC-ICRS/16-17/03]. The
research of SBL was supported by the National Natural Science
Foundation of China [Grant No. 61876133], and the research of DXZ
was partially supported by the Research Grant Council of Hong Kong [Project No. {  CityU 11306318}]}}

\author{Charles K. Chui$^{1,2}$, Shao-Bo Lin$^{3,4}$\thanks{Corresponding author: sblin1983@gmail.com}, Ding-Xuan Zhou$^{4}$  }

  \date{}
\maketitle

\begin{center}

 \footnotesize 
1. Department of Mathematics, Hong Kong Baptist University, { Hong Kong}

2. Department of Statistics,  Stanford University,
CA 94305, U.S.A.

3.  Department of Mathematics, Wenzhou University, Wenzhou,
China

4. School of Data Science and Department of Mathematics, City University of Hong Kong, {  Hong Kong}

\begin{abstract}

Based on the tree architecture, the objective of this paper is to
design deep neural networks with two or more hidden layers (called
deep nets) for realization of radial functions so as to enable
rotational invariance for near-optimal function approximation in an
arbitrarily high dimensional Euclidian space. It is shown that deep
nets have much better performance than shallow nets (with only one
hidden layer) in terms of approximation accuracy and learning
capabilities. In particular, for learning radial functions, it is
shown that near-optimal rate can be achieved by deep nets but not by
shallow nets. Our results illustrate the necessity of depth in
neural network design for realization of rotation-invariance target
functions.

{\bf Keywords:}  Deep nets, rotation-invariance, learning theory , radial-basis functions\\
\end{abstract}
\end{center}

\section{Introduction}
In this era of big data, data-sets of massive size and with various
features are routinely acquired, creating a crucial challenge to
machine learning in the design of learning strategies for data
management, particularly in realization of certain data features.
Deep learning \cite{Hinton2006} is a state-of-the-art approach for
the purpose of realizing such features, including localized position
information \cite{Chui1994,Chui2017}, geometric structures of
data-sets \cite{Chui2016,Shaham2015}, and data sparsity
\cite{Linh2017,Lin2018a}. For this and other reasons, deep learning
has recently received much attention, and has been successful  in
various application domains \cite{Goodfellow}, such as computer
vision,  speech recognition, image classification, fingerprint
recognition and earthquake forecasting.

Affine transformation-invariance, and particularly
rotation-invariance, is an important data feature, prevalent in such
areas as statistical physics \cite{Linh2017}, early warning of
earthquakes \cite{Satriano2011}, 3-D point-cloud segmentation
\cite{Qi2017}, and image rendering \cite{Meylan2006}. Theoretically,
neural networks with one hidden layer (to be called shallow nets)
are incapable of embodying rotation-invariance features in the sense
that its performance in handling these features is analogous to the
failure of algebraic polynomials \cite{Konovalov2008} in handling
this task \cite{Konovalov2009}. The primary goal of this paper is to
construct neural networks with at least two hidden layers (called
deep nets) to realize rotation-invariant features by deriving
``fast" approximation and learning rates of radial functions as
target functions.

Recall that a function $f$ defined on the $d-$dimensional ball,
$\mathbb B^d(R)$ with radius $R>0$ where $d\ge2$, is called a radial
function, if there exists a univariate real-valued function $g$
defined on the interval $[0, R]$ such that $f({\bf x}) = g(|{\bf
x}|^2)$, for all ${\bf x} \in \mathbb B^d(R)$. For convenience, we
allow $\mathbb B^d(R)$ to include the Euclidian space $\mathbb R^d$
with $R=\infty$. Hence, all radial-basis
functions (RBF's) are special cases of radial functions. In this
regard, it is worthwhile to mention that the most commonly used
RBF's are the multiquadric $g(r) = (r^2 + c)^{1/2}$ and Gaussian
$g(r) = e^{-c r^2}$, where $c>0$. For these and some other RBF's,
existence and uniqueness of scattered data interpolation from the
linear span of $\{f({\bf x} - {\bf x}_k): k = 1, \cdots, {  \ell}\}$, for
arbitrary distinct centers $\{{\bf x}_1, \cdots, {\bf x}_{ \ell}\}$ and
for any ${  \ell\in\mathbb N}$, are assured. The reason for the popularity of the
multiquadric RBF is fast convergence rates of the interpolants to the
target function \cite{Buhmann2003}, and that of the Gaussian RBF is
that it is commonly used as the activation function for constructing
radial networks that possess the universal approximation property
and other useful features {  (see \cite{McCane2017}, \cite{Mhaskar2004},
\cite{Ye2008}, \cite{Zhou2018}, \cite{Zhou2018a}, \cite{Guo2019})} and references therein). The departure of our paper
from constructing radial networks is that since RBF's are radial
functions, they qualify to be target functions for our
general-purpose deep nets with general activation functions. Hence,
if the centers $\{{\bf x}_1, \cdots, {\bf x}_{\red \ell}\}$ of the desired RBF
have been chosen and the coefficients $a_1, \cdots, a_{ \ell}$ have been
pre-computed, then the target function
$$ \sum_{k=1}^{{  \ell}}a_k f({\bf x} - {\bf x}_k)$$ can be realized by using
one extra hidden layer for the standard arithmetic operations of
additions and multiplications and an additional outer layer for the
input of RBF centers and coefficients to the deep net constructed in
this paper.

The main results of this paper are three-fold. We will first derive
a lower bound estimate for approximating radial functions by deep
nets. We will then construct a deep net with four hidden layers to
achieve this lower bound (up to a logarithmic multiplicative factor)
to illustrate the power of depth in realizing rotation-invariance.
Finally, based on the prominent approximation ability of deep nets,
we will show that implementation of the empirical risk minimization
(ERM) algorithm in deep nets facilitates fast learning rates  and is
independent of dimensions. The presentation of this paper is
organized as follows. Main results will be stated in Section
\ref{Sec. Main results}, where near-optimal approximation order and
learning rate of deep nets are established.  In Section
\ref{Sec.Tools}, we will establish our main tools for constructing
deep nets with two hidden layers for approximation of univariate
smooth functions. Proofs of the main results will be provided in
Section \ref{Sec.Proof}. Finally, derivations of the auxiliary
lemmas that are needed for our proof of the main results are
presented in Section \ref{Sec. Auxiliary Lemmas}.

\section{Main Results}\label{Sec. Main results}
Let $\mathbb B^d := \mathbb B^d(1)$  denote the unit ball in
$\mathbb R^d$ with center at the origin.  Then any radial function
$f$ defined on $\mathbb B^d$ is represented by $f({\bf x})=g(|{\bf
x}|^2)$ for some function $g:[0,1]\rightarrow\mathbb R$. Here and
throughout the paper, the standard notation of the Euclidean norm
$|{\bf x}|:=[(x^{(1)})^2+\dots+(x^{(d)})^2]^{1/2}$ is used for ${\bf
x}:=(x^{(1)},\dots,x^{(d)})\in\mathbb R^d$. In this section, we
present the main results on approximation and learning of radial
functions $f$.

\subsection{Deep nets with tree structure}
Consider the collection
\begin{equation}\label{shallow net}
       \mathcal S_{\phi,n}:=\left\{\sum_{j=1}^na_j\phi({\bf w}_j\cdot
       {\bf x}+b_j): a_j,b_j\in\mathbb R, {\bf w}_j\in\mathbb R^d\right\},
\end{equation}
of shallow nets with activation function $\phi:\mathbb
R\rightarrow\mathbb R$, where ${\bf x}\in\mathbb B^d$. The deep nets
considered in this paper are defined recursively in terms of shallow
nets according to the tree structure, as follows:
\begin{definition}\label{Definition: neural networks}
Let $L,N_1,\dots, N_L\in\mathbb N$, $N_0=d$, and {  $\phi_k:\mathbb
R\rightarrow\mathbb R$, $k=0,1,\dots,L$,} be univariate activation
functions. Set {
$$
       H_{\vec{\alpha},0}({\bf x})=\sum_{j=1}^{N_0}a_{j,\vec{\alpha},0}\phi_0(w_{j,\vec{\alpha},0}x^{(j)}+b_{j,\vec{\alpha}_0,0}),\qquad {\bf x}=(x^{(1)},\dots,x^{(d)}),\vec{\alpha}\in
       \prod_{i=1}^L\{1,2,\dots,N_i\}.
$$
Then a deep net with the tree structure of $L$ layers can be
formulated recursively by {
$$
   H_{\vec{\alpha},k}({\bf x})=\sum_{j=1}^{N_{k}}a_{j,\vec{\alpha},k}\phi_{k}(H_{j,\vec{\alpha},k-1}({\bf x})+b_{j,\vec{\alpha},k}), \qquad\ 1\leq k\leq L, \vec{\alpha}\in
   \prod_{i=k+1}^L\{1,2,\dots,N_i\},
$$
where $a_{j,\vec{\alpha},k},b_{j,\vec{\alpha},k},w_{j,\vec{\alpha},0}\in\mathbb R$  for each
$j\in\{1,2,\dots,N_k\}$, $\vec{\alpha}\in \prod_{i=k+1}^L\{1,2,\dots,N_i\}$, and $k\in\{0,1,\dots,L\}$. Let $\mathcal
H^{tree}_L$ denote the set of output functions $H_{L}=H_{\vec{\alpha},L}$ for $\vec{\alpha}\in\varnothing$ at the $L$-th
layer.}}
\end{definition}

{
\begin{figure}[H]
\centering
\includegraphics*[scale=0.4]{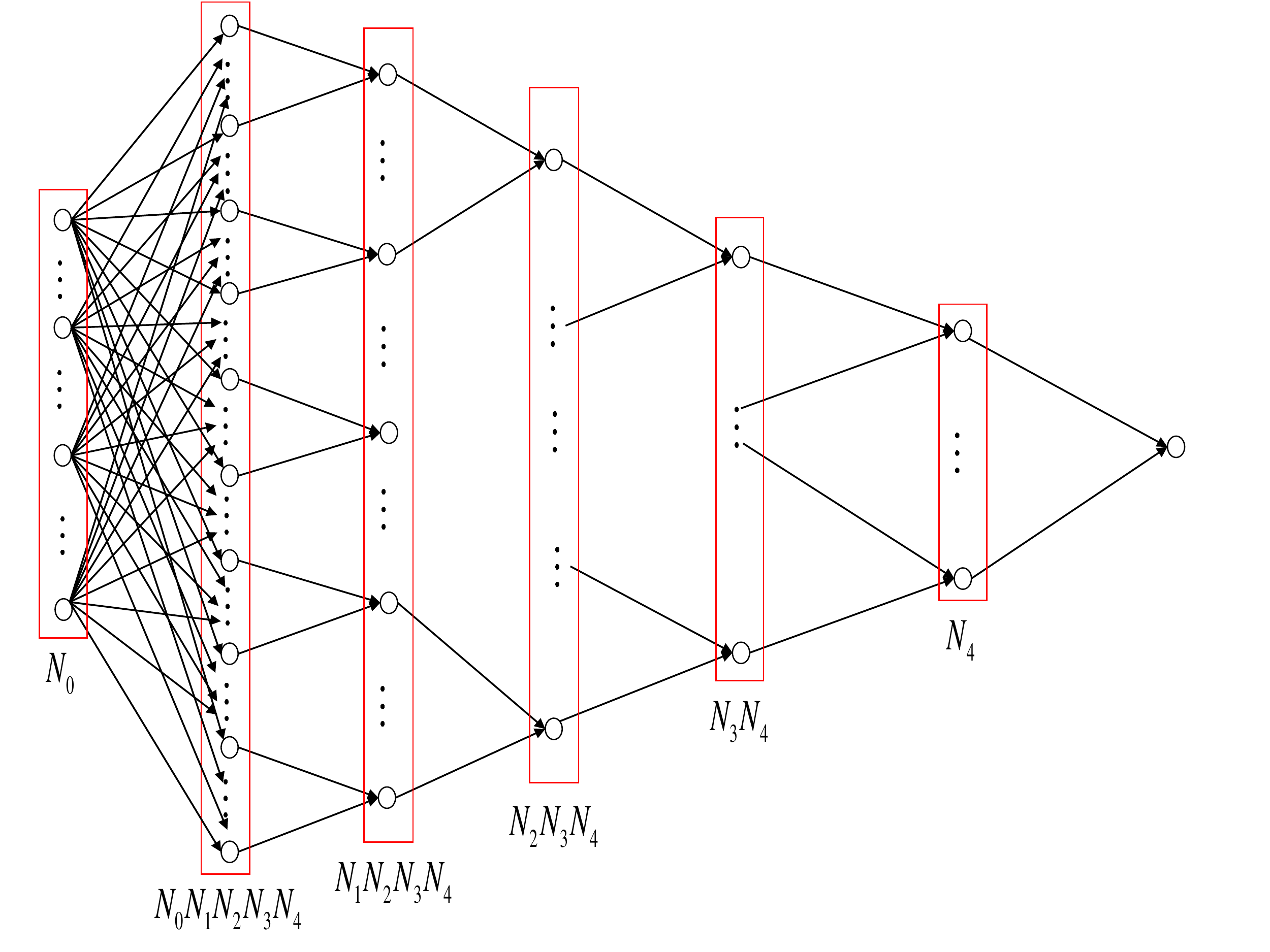}
 \label{Fig:Structure}
 \caption{Tree structure of  deep nets with six layers}
\end{figure}
}

Note that if the initial activation function is chosen to be
$\phi_0(t)=t$ and ${  b_{j,\alpha,0}=0}$, then $\mathcal H_1^{tree}$ is the
same as the shallow net $\mathcal S_{\phi_1,N_1}$. Figure
\ref{Fig:Structure}  exhibits the structure of the deep net defined
in Definition \ref{Definition: neural networks}, showing sparse and
tree-based connections among neurons. Due to the concise
mathematical formulation, this definition of deep nets \cite{Chui2019} has been
widely used to illustrate its advantages over shallow nets. In
particular, it was shown in \cite{Mhaskar1993} that deep nets with
the tree structure can be constructed to overcome the saturation
phenomenon of shallow nets; in \cite{Maiorov1999b} that {    deep
nets}, with two hidden layers, tree structure, and finitely many
neurons, can be constructed to possess the universal approximation
property; and in \cite{Mhaskar2016a, Kohler2017} that deep nets with
the tree structure are capable of embodying tree structures for data
management. In addition, a deep net with the tree structure was
constructed in \cite{Chui2017}  to realize manifold data.

As a result of the sparse connections of deep nets with the tree
structure, it  follows from Definition \ref{Definition: neural
networks} and Figure 1 that there are a total of
\begin{equation}\label{Def:AL}
      \mathcal A_L:=2\sum_{k=0}^L\Pi_{\ell=0}^{L-k}N_{L-\ell}+ \Pi_{\ell=0}^{L}N_{\ell}
\end{equation}
free parameters for   $H_L\in\mathcal H_L^{tree}$. For
$\alpha,\mathcal R\geq 1$, we introduce the notation {
\begin{eqnarray}\label{deep nets class}
      \mathcal H^{tree}_{L,\alpha,\mathcal R}&:=&\large\{ H_{L}\in\mathcal
      H^{tree}_L:|a_{{j,\vec{\alpha},k}}|,|b_{j,\vec{\alpha},k}|,|w_{j,\vec{\alpha},0}|\leq \mathcal
      R \left(\mathcal A_L\right)^\alpha,\nonumber\\
      &&
      0\leq k\leq L,1\leq j\leq N_k,\vec{\alpha}\in\prod_{i=k+1}^L\{1,2,\dots,N_i\} \large\}.
\end{eqnarray}
}
{  For functions in this class,   the parameters of
deep nets are bounded.} This is indeed a necessity condition, since
it follows from the results in \cite{Maiorov1999b, Maiorov1999c}
there exists some $h\in\mathcal H^{tree}_{2,\infty,\infty}$ with
finitely many neurons but infinite capacity (pseudo-dimension). The
objective of this paper is to construct deep nets of the form
(\ref{deep nets class}) for some $\alpha$ and $\mathcal R$, for the
purpose of approximating and learning radial functions.

\subsection{Lower bounds for approximation by deep nets }

In this subsection, we show the power of depth in approximating
radial functions, by showing some lower bound results for
approximation by deep nets under certain smoothness assumption on
the radial functions.

\begin{definition}\label{Definition:smoothness}
For $\mathbb A\subset\mathbb R$, $c_0>0$ and $r=s+v$, with
$s\in\mathbb N_0:=\{0\}\cup\mathbb N$ and $0<v\leq 1$, let
$Lip^{(r,c_0)}_{\mathbb A}$ denote the collection of univariate
$s$-times differentiable functions $g:\mathbb A\rightarrow\mathbb
R$, whose $s$-th derivatives satisfy the Lipschitz condition
\begin{equation}\label{lip}
          |g^{(s)}(t)-g^{(s)}(t_0)|\leq c_0|t-t_0|^v,\qquad\forall\
          t,t_0\in\mathbb A.
\end{equation}
In particular, for $\mathbb A = \mathbb I:=[0,1]$, let
$Lip^{(\diamond,r,c_0)}$ denote the set of radial functions $f(\bf
x)=g(|{\bf x}|^2)$ with $g\in Lip^{(r,c_0)}_{\mathbb I}$.
\end{definition}

We point out that the above Lipschitz continuous assumption is
standard for radial basis functions (RBF's) in Approximation Theory,
and was adopted in \cite{Konovalov2008,Konovalov2009} to quantify
the approximation abilities of polynomials and ridge functions. For
$U,V\subseteq L_p(\mathbb B^d)$ and $1\leq p\leq\infty$, we denote
by
$$
     \mbox{dist}(U,V, L_p(\mathbb B^d)):=\sup_{f\in U} \mbox{dist}(f,V, L_p(\mathbb B^d))
     :=\sup_{f\in U} \inf_{{ g}\in V}\|f-{  g}\|_{L_p(\mathbb
       B^d)}
$$
the deviations of $U$ from $V$ in  $L_p(\mathbb B^d)$. The following
main result shows that shallow nets are incapable of embodying the
rotation-invariance property.

\begin{theorem}\label{Theorem:lower bound for deep nets 1}
Let $d\geq 2$, $n,L\in \mathbb N$, $c_1>0$, $\mathcal R,\alpha\geq1$
and $\mathcal H^{tree}_{L,\alpha,\mathcal R}$ be defined by
(\ref{deep nets class}) with ${  \tilde{n}=\mathcal A_L}$ free parameters, and
$\mathcal A_L$ be given by (\ref{Def:AL}). Suppose that $\phi_j\in
Lip_{\mathbb R}^{(1,c_1)}$ satisfies $\|\phi_j\|_{L_\infty(\mathbb
R)}\leq 1$ for every $j\in\{0,1,\dots,L\}$. Then { for $c_0>0$, $r=s+v$ with $s\in\mathbb N_0$ and $0<v\leq 1$,}
\begin{equation}\label{limitation for shallow}
   \mbox{dist}(Lip^{(\diamond,r,c_0)},\mathcal S_{\phi_1,n},L_\infty(\mathbb B^d))
   \geq
    C^*_1{(d+2)n}^{-r/(d-1)},
\end{equation}
and
\begin{equation}\label{lower bound for deep1}
   \mbox{dist}(Lip^{(\diamond,r,c_0)},\mathcal H^{tree}_{L,\alpha,\mathcal R},L_\infty(\mathbb B^d))\geq
  C_2^*(L^2{ \tilde{n}}\log_2 { \tilde{n}})^{-r},\qquad L\geq 2,
\end{equation}
where $(d+2)n$ is the number of parameters for the shallow net ${\mathcal S}_{\phi_1, n}$ and the constants $C_1^*$ and $C_2^*$ are independent of $n$, $\tilde{n}$ or
$L$.
\end{theorem}

The proof of Theorem \ref{Theorem:lower bound for deep nets 1} is
postponed to Section \ref{Sec.Proof}. Observe that Theorem
\ref{Theorem:lower bound for deep nets 1} exhibits an interesting
phenomenon in approximation of radial functions by deep nets, in
that  the depth plays a crucial role, by comparing (\ref{limitation
for shallow}) with (\ref{lower bound for deep1}). For instance, the
lower bound  $({ \tilde{n}}\log { \tilde{n}})^{-r}$ for deep nets is a big improvement of
the lower bound ${ \tilde{n}}^{-r/(d-1)}$ for shallow nets,  for dimensions $d
>2$.

\subsection{Near-optimal approximation rates for deep nets}
In this subsection, we  show  that the {  lower bound (\ref{lower bound for deep1})} is achievable up
to a logarithmic factor by some deep net with $L=3$ layers for
certain commonly used activation functions that satisfy the
following smoothness condition.

\begin{assumption}\label{Assumption:smooth} The activation function $\phi$ is assumed to be infinitely differentiable, with both $\|\phi'\|_{L_\infty(\mathbb R)}$ and
$\|\phi\|_{L_\infty(\mathbb R)}$ bounded by $1$, such that
$\phi^{(j)}(\theta_0)\neq0$ for some $\theta_0\in \mathbb R$  and
all ${  j\in\mathbb N_0,} $ and that
\begin{equation}\label{sigmoid rate}
     |\phi(-t)|=\mathcal O(t^{-1}),\qquad |1-\phi(t)|=\mathcal
     O(t^{-1}),\qquad t\rightarrow\infty.
\end{equation}
\end{assumption}
It is easy to see that all of the logistic function:
$\phi(t)=\frac{1}{1+e^{-t}}$, the hyperbolic tangent function:
$\phi(t)=\frac12(\tanh(t)+1)$, the arctan function:
$\phi(t)=\frac1{\pi}\arctan(t)+\frac12$,  and the Gompertz function:
$\phi(t)=e^{-e^{-t}}$, satisfy Assumption \ref{Assumption:smooth},
in which we essentially impose three conditions on the activation
function $\phi$, namely: infinite differentiability, non-vanishing
of all derivatives at the same point, and the sigmoidal property
(\ref{sigmoid rate}). On the other hand, we should point out that
such strong assumptions are stated only for the sake of brevity, but
can be relaxed to Assumption \ref{Assumption:smooth1} in Section
\ref{Sec.Tools} below. In particular, the infinite differentiability
condition on $\phi$ can be replaced by some much weaker smoothness
property as that of the target function $f$. The following is our
second main result, which shows that deep nets can be constructed to
realize the rotation-invariance property of $f$ by exhibiting a
dimension-independent approximation error bound, which is much
smaller than that for shallow nets.

\begin{theorem}\label{Theorem:almost optimal}
Let $n\geq 2$, {  $c_0>0$}, and $r=s+v$ with $s\in\mathbb N_0$ and $0<v\leq 1$.
Then under Assumption \ref{Assumption:smooth}, {  for $\mathcal R,\alpha\geq 1$},
\begin{equation}\label{Optimal approx}
  9^{-r}C_2^*(n\log n)^{-r}\leq\mbox{dist}(Lip^{(\diamond,r,c_0)},\mathcal H^{tree}_{3,\alpha,\mathcal R},
   L_\infty(\mathbb B^d))\leq C_3^*n^{-r},
\end{equation}
where $\mathcal H^{tree}_{3,\alpha,\mathcal R}$ is defined by
(\ref{deep nets class}) with $L=3$, $N_0=d,N_1=6,N_2=s+3,N_3=3n+3$,
{  $\alpha=48(3+r(r+1)+ r(s+1)!7(r+1))$}, and {  the constant  $C_3^*$ is independent of $n$.}
\end{theorem}
{ Note that the deep net in Theorem \ref{Theorem:almost optimal} has the number of free parameters satisfying
$$
     6d(s+3) (3n+3)\leq \tilde{n}=\mathcal A_3\leq 54d(s+3) (3n+3).
$$
}
It follows from (\ref{Optimal approx}) that, up to a logarithmic
factor, there exists a deep net with $L=3$ and some commonly used
activation functions that achieves the lower bound { (\ref{lower bound for deep1})} established in
Theorem \ref{Theorem:lower bound for deep nets 1}.

We would like to mention an earlier work \cite{McCane2017} on
approximating radial functions by deep ReLU networks, where it was
shown that for each $f\in Lip^{(\diamond,1,c_0)}$, there exists a
fully connected deep  net $H^{ReLU}_{ \tilde{n}}$ with ReLU activation
function, $\phi(t)=\max\{t,0\}$, and  at least ${ \tilde{n}}$ parameters
 and at least {  $\mathcal O(\log\tilde{n})$} layers, such
that {
$$
    \|f-H^{ReLU}_{ \tilde{n}}\|_{L^\infty(\mathbb B^d)}\leq C_4^*
    { \tilde{n}}^{-\frac{1}{2\jmath}}
$$}
for some absolute constant ${  \jmath\geq 1}$ and  constant $C_4^*$ independent of ${ \tilde{n}}$. The novelties of our
results in the present paper, as compared with those in
\cite{McCane2017}, can be summarized as follows. Firstly, noting
that  ${ \tilde{n}}^{-\frac{1}{2\jmath}}\gg ({ \tilde{n}}\log { \tilde{n}})^{-1}$ for $\jmath\geq 1$,
we may conclude that only an upper bound (without approximation
order estimation) was provided in \cite{McCane2017}, while both
near-optimal approximation error estimates and achievable lower
bounds are derived in our present paper on the approximation of
functions in $Lip^{(\diamond,r,c_0)}$. In addition, while fully
connected deep nets were considered in \cite{McCane2017}, we
construct a deep net with sparse connectivity in our paper. Finally,
to achieve upper bounds for any $r>0$ (as opposed to merely $r=1$),
non-trivial techniques, such as ``product-gate''   and approximation
of smoothness functions by products of deep nets and Taylor
polynomials are introduced in Section \ref{Sec.Tools}. It would {  be
of  interest} to obtain similar results as Theorem
\ref{Theorem:almost optimal} for deep ReLU nets, but this is not
considered in the present paper.

\subsection{Learning rate analysis for empirical risk minimization on deep nets}

Based on near-optimal approximation error estimates in Theorem
\ref{Theorem:almost optimal}, we shall deduce a near-optimal
learning rate for the algorithm of empirical risk minimization (ERM)
over $\mathcal H^{tree}_{3,\alpha,\mathcal R}$. Our analysis will be
carried out in the standard regression framework \cite{Cucker2007},
with samples $D_m=\{(x_i,y_i)\}_{i=1}^m$ drawn independently
according to an unknown Borel {  probability measure} $\rho$ on ${\mathcal
Z}={\mathcal X}\times {\mathcal Y}$, with $\mathcal X=\mathbb B^d$
and $\mathcal Y\subseteq[-M,M]$ for some $M>0$.

The primary objective is to {  learn} the regression function
$f_\rho(x)=\int_{\mathcal Y} y d\rho(y|x)$ that minimizes the
generalization error $\mathcal E(f):=\int_{\mathcal
Z}(f(x)-y)^2d\rho,$ where $\rho(y|x)$ denotes the conditional
distribution at $x$ induced by $\rho$. To do so, we consider the
learning rate for the {  ERM algorithm
\begin{equation}\label{ERM}
      f_{D,n,\phi}:=\arg\min_{f\in\mathcal H^{tree}_{3,\alpha,\mathcal R}}\frac1m\sum_{i=1}^m\left(f(x_i)-y_i\right)^2.
\end{equation}
Here, $n\in\mathbb N$ is the parameter appearing in the definition of $\mathcal H^{tree}_{3,\alpha,\mathcal R}$. Since} $|y_i|\leq M$, it is natural to project the final output
$f_{D,n,\phi}$ to the interval $[-M, M]$ by the truncation operator
$\pi_M
f_{D,n,\phi}(x):=sign(f_{D,n,\phi}(x))\min\{|f_{D,n,\phi}(x)|,M\}.$
The following theorem is our third main result on a near-optimal
dimension-independent learning rate for $\pi_M f_{D,n,\phi}$.

{
\begin{theorem}\label{Theorem: ERM}
Let  $f_{D,n,\phi}$ be defined by (\ref{ERM}), and
consider  $f_\rho\in Lip^{(\diamond,r,c_0)}$ with $c_0>0$ and $r=s+v$ with $s\in\mathbb N_0$, $0<v\leq 1$,  and
$n=\left[C_5^*m^{\frac{1}{2r+1}}\right]$. Then under Assumption
\ref{Assumption:smooth}, for any $0<\delta<1$,
\begin{equation}\label{learning rate}
      \mathcal E(\pi_Mf_{D,n,\phi})-\mathcal E(f_\rho)
      \leq
      C_6^*m^{-\frac{2r}{2r+1}}\log (m+1)\log\frac3\delta
\end{equation}
holds  with confidence   at least $1-\delta$. Furthermore,
\begin{eqnarray}\label{almost optimal learning rate}
          C_7^* m^{-\frac{2r}{2r+1}}
            \leq
         \sup_{f_\rho\in Lip^{(\diamond,r,c_0)}}E\left\{\mathcal E(\pi_Mf_{D,n,\phi})-\mathcal
         E(f_\rho)\right\}
          \leq
          C_8^* m^{-\frac{2r}{2r+1}}\log (m+1),
\end{eqnarray}
where, as usual, $[a]$  denotes the integer part of $a>0$ and the
constants  $C_5^*,C_6^*,C_7^*,C_8^*$  are  independent of $\delta$, $m$
and $n$.
\end{theorem}
}

We emphasize that the learning rate in (\ref{learning rate}) is
independent of the dimension $d$, and is much better than the
optimal learning rate $m^{-\frac{2r}{2r+d}}$ for learning
$(r,c_0)$-smooth (but not necessarily radial) functions on $\mathbb
B^d$ \cite{Gyorfi2002,Lin2016,Lin2018}.  For shallow nets, it
follows from (\ref{limitation for shallow}) that to achieve a
learning rate similar to (\ref{almost optimal learning rate}), we
need at least $[m^{\frac{d-1}{2r+1}}]$ neurons to guarantee the
$\mathcal O(m^{-\frac{2r}{2r+1}})$ bias. For $d\geq 3$, since
$m^{\frac{d-1}{2r+1}}\geq m^{\frac{1}{2r+1}}$, the capacity of
neural networks is large. Consequently, it is difficult to derive a
satisfactory variance, so that  derivation of a similar almost
optimal learning rates as (\ref{almost optimal learning rate}) for
ERM on shallow nets is also difficult. {  Thus,
Theorem \ref{Theorem: ERM} demonstrates } that ERM on deep nets can embody the
rotation-invariance property by deducing the learning rate of order
$m^{-\frac{2r}{2r+1}}$.

\section{Approximation by Deep Nets without Saturation}\label{Sec.Tools}

Construction of  neural networks  to approximate smooth functions is
a classical and long-standing topic in approximation theory.
Generally speaking, there are two approaches, {  one by}
constructing neural networks to approximate algebraic polynomials,
and the other by constructing neural networks with localized
approximation properties. The former usually requires extremely
large {  norms of weights} \cite{Mhaskar1996,Xie2010} and the latter
frequently suffers from the well-known saturation phenomenon
\cite{Chen1993, Chui1994}, in the sense that the approximation rate
cannot be improved any further, when the regularity of the target
function goes beyond a specific level. The novelty of our {  method  is} to adopt the ideas from both of the above two
approaches to construct a deep net with two hidden layers with
controllable {  norms} of weights and without saturation, by
considering the ``exchange-invariance'' between polynomials and
shallow nets, the localized approximation of neural networks, a
recently developed ``product-gate'' technique
\cite{Yarotsky2017}, and a novel Taylor formula.  For this purpose,
we need to impose differentiability and the sigmoid property on
activation functions, as follows.

\begin{assumption}\label{Assumption:smooth1}
Let {  $c_0>0$,} $r_0=s_0+v_0$ with $s_0\geq2$ and $0<v_0\leq 1$. Assume that
$\phi\in Lip^{(r_0,c_0)}_{\mathbb R}$ is a sigmoidal function with
$\|\phi'\|_{L_\infty(\mathbb R)},\|\phi\|_{L_\infty(\mathbb
R)}\leq1$, such that $\phi^{(j)}(\theta_0)\neq0$ for all
$j=0,1,\dots,s_0$, for some $\theta_0\in \mathbb R$.
\end{assumption}

It is obvious that Assumption \ref{Assumption:smooth1} is much
weaker than the smoothness property of $\phi$ in Assumption
\ref{Assumption:smooth}.  Furthermore, it removes the restriction
(\ref{sigmoid rate}) on the use of sigmoid functions as activation
function, by considering only the general sigmoidal property:
$$
      \phi(-t) \rightarrow0, \qquad\mbox{and}\quad \phi(t)\rightarrow 1,\qquad \mbox{when} \quad t\rightarrow\infty.
$$
In view of this property, we introduce the notation
\begin{equation}\label{def delta111}
                      \delta_\phi(A):=\sup_{t\geq
                      A}\max(|1-\phi(t)|,|\phi(-t)|),
\end{equation}
where $A\geq 1$, and observe that
$\lim_{A\rightarrow\infty}\delta_\phi(A)=0$.
\subsection{Exchange-invariance of univariate polynomials and shallow nets}

In this subsection, a shallow net with one neuron    is constructed to
replace {  a}  univariate homogeneous polynomial {  together with a polynomial of lower degree}.  It is shown in the
following proposition that such a replacement does not degrade the
polynomial approximation property.

\begin{proposition}\label{Proposition:poly for nn plus poly}
Under Assumption \ref{Assumption:smooth1} with {  $c_0>0$}, $r_0=s_0+v_0$ and
$\theta_0\in\mathbb R$, let $k\in\{0,\dots,s_0\}$ and
$p_k(t)=\sum_{i=0}^ku_{i}t^i$ with $u_k\neq0$. Then for an arbitrary
$\varepsilon\in(0,1)$,
\begin{equation}\label{error estimate for p bb n p}
       \left|p_k(t)-u_{k}\frac{k!}{\mu_k^k\phi^{(k)}(\theta_0)}\phi(\mu_kt+\theta_0) - p^*_{k-1}(t)\right|\leq  \varepsilon, \qquad \forall\ t\in [-1,1],
\end{equation}
where
\begin{equation}\label{muk}
  \mu_k:=\mu_{k,\varepsilon}:=\left\{\begin{array}{cc}
     \min\left\{1,\frac{\varepsilon|\phi^{(k)}(\theta_0)|(k+1)}{|u_{k}|
       \max_{\theta_0-1\leq t\leq
          \theta_0+1}|\phi^{(k+1)}(t)|}\right\}, &  \mbox{if}\ 0\leq k\leq s_0-1\\
           \min\left\{1,\left[\frac{\varepsilon|\phi^{(s_0)}(\theta_0)|\Gamma(s_0+v_0+1)}{s_0!\Gamma(v_0+1)c_0|u_{s_0}|
       }\right]^\frac1{v_0}\right\}, &  \mbox{if}\ k=s_0,
     \end{array}\right.
\end{equation}
$p^*_{-1}(t)=0$ and
\begin{eqnarray}\label{P(k-1)}
                p^*_{k-1}(t)
                :=\sum_{i=0}^{k-1}u^*_{i}t^i
                :=
                \sum_{i=0}^{k-1}\left(u_{i}-\frac{u_{k}k!\phi^{(i)}(\theta_0)}{\phi^{(k)}(\theta_0)\mu_k^{k-i}i!}\right)t^i.
\end{eqnarray}
\end{proposition}

The proof of Proposition \ref{Proposition:poly for nn plus poly}
requires the following {  Taylor   representation}  which is an
easy consequence of the classical Taylor formula
$$
           \psi(t)=\sum_{i=0}^{\ell-1}\frac{\psi^{(i)}(t_0)}{i!}(t-t_0)+\frac1{(\ell-1)!}
           \int_{t_0}^t\psi^{(\ell)}(u)(t-u)^{(\ell-1)}du
$$
with remainder in integral form, {  and} using the formula
$\int_{t_0}^t(t-u)^{\ell-1}du=\frac{(t-t_0)^{\ell}}\ell.$ To obtain
the Taylor polynomial of degree $k$,  this formula does not require
$\psi$ to be $(k+1)$-times differentiable. This observation is
important throughout our analysis.

\begin{lemma}\label{Lemma:Taylor formula} Let $\ell\geq1$ and $\psi$ be $\ell$-times
differentiable {  on $\mathbb R$}.Then {  for}   $t,t_0\in \mathbb R$,
\begin{equation}\label{Taylor formula1}
           \psi(t)=\psi( t_0)+\frac{\psi'(t_0)}{1!}
            (t-t_0)+\cdots+\frac{\psi^{(\ell)}(t_0)}{\ell!}( t-t_0)^\ell+r_\ell(t),
\end{equation}
where
\begin{equation}\label{def for r1}
            r_\ell(t)=\frac{1}{(\ell-1)!}\int_{t_0}^t\left[\psi^{(\ell)}(u)-\psi^{(\ell)}(t_0)\right](t-u)^{\ell-1}du.
\end{equation}
\end{lemma}

We are now ready to prove Proposition \ref{Proposition:poly for nn
plus poly}.

{\bf Proof of Proposition \ref{Proposition:poly for nn plus
poly}.} Since  $\mu_k\in(0,1]$ from its definition, we may apply
  Lemma \ref{Lemma:Taylor formula} with $t_0=\theta_0$ and
$\ell=k$ to obtain
$$
           \phi(\mu_kt+\theta_0)=\sum_{i=0}^{k}\frac{\phi^{(i)}(\theta_0)}{i!}(\mu_kt)^i+r_{k,\mu_k}(t),
$$
where $r_{0,\mu_0}=\phi(\mu_kt+\theta_0)-\phi(\theta_0)$ and
\begin{equation}\label{rk}
            r_{k,\mu_k}(t):=\frac{1}{(k-1)!}\int_{\theta_0}^{\mu_kt+\theta_0}
            \left[\phi^{(k)}(u)-\phi^{(k)}(\theta_0)\right](\mu_kt+\theta_0-u)^{k-1}du
\end{equation}
for $k\geq 1$. {  It follows} that
$$
             t^k=\frac{k!}{\mu_k^k\phi^{(k)}(\theta_0)}\phi(\mu_k
              t+\theta_0)+q_{k-1}(t)-\frac{k!}{\mu_k^k\phi^{(k)}(\theta_0)}r_{k,\mu_k}(t),
$$
where
$$
    q_{k-1}(t)=\frac{-k!}{\mu_k^k\phi^{(k)}(\theta_0)}\sum_{i=0}^{k-1}\frac{\phi^{(i)}(\theta_0)}{i!}(\mu_kt)^i,
$$
so that
\begin{eqnarray*}\label{po1aaa}
             p_k(t)
        =
     u_{k}\frac{k!}{\mu_k^k\phi^{(k)}(\theta_0)}\phi(\mu_kt+\theta_0)
     +p^*_{k-1}(t)-u_{k}\frac{k!}{\mu_k^k\phi^{(k)}(\theta_0)}r_{k,\mu_k}
              (t),
\end{eqnarray*}
 with $p^*_{k-1}$ defined by (\ref{P(k-1)}). What is left is to estimate the remainder  $u_{k}\frac{k!}{\mu_k^k\phi^{(k)}(\theta_0)}r_{k,\mu_k}(t)$. To this end,  we observe, {  for the case $k=0$ , from the definition of $\mu_0$, that for any $t\in[-1,1]$,
$$
   \left|u_{0}\frac{1}{\phi(\theta_0)}r_{0,\mu_0}(t)\right|
   \leq\frac{|u_0|}{|\phi(\theta_0)|}\max_{\theta_0-1\leq\tau\leq\theta_0+1}|\phi'(\tau)|\mu_0|t|
   \leq\frac1{|\phi(\theta_0)|}\varepsilon|\phi(\theta_0)|=\varepsilon.
$$}
For $1\leq k\leq s_0-1$, we may apply the estimate
$$
    \left|\phi^{(k)}(\mu_ku+\theta_0)-\phi^{(k)}(\theta_0)\right|\leq  \max_{\theta_0-1\leq\tau\leq\theta_0+1}
    |\phi^{(k+1)}(\tau)| \mu_k|u|,\qquad \forall\ u\in[0,t],t\in[-1,1]
 $$
to compute, for any $t\in[-1,1],$
\begin{eqnarray*}
    &&\left|u_{k}\frac{k!}{\mu_k^k\phi^{(k)}(\theta_0)}r_{k,\mu_k}(t)\right|
    =\left|\frac{ku_k}{\phi^{(k)}(\theta_0)}\int_0^t[\phi^{(k)}(\mu_ku+\theta_0)-\phi^{(k)}(\theta_0)]
    (t-u)^{k-1}du\right|\\
    &\leq &
    k(k+1)\varepsilon\int_0^1u(1-u)^{k-1}du
    =k(k+1)\varepsilon\frac{\Gamma(2)\Gamma(k)}{\Gamma(k+2)}=\varepsilon.
 \end{eqnarray*}
 Finally, for $k=s_0$, we may apply the Lipschitz property of $\phi^{(s_0)}$ to obtain
$$
    \phi^{(s_0)}(\mu_ku+\theta_0)-\phi^{(s_0)}(\theta_0)\leq  c_0|\mu_ku|^{v_0},\qquad\forall\ {  u\in[0,t],t\in[-1,1],}
$$
so  that for any $t\in[-1,1]$, we have
 \begin{eqnarray*}
    &&\left|u_{s_0}\frac{s_0!}{\mu_{s_0}^{s_0}\phi^{(s_0)}(\theta_0)}r_{s_0,\mu_{s_0}}(t)\right|
      =
      \left| \frac{s_0u_{s_0}}{\phi^{(s_0)}(\theta_0)}\int_{0}^{t}
            \left[\phi^{(s_0)}(\mu_{s_0}u+\theta_0)-\phi^{(s_0)}(\theta_0)\right](t-u)^{s_0-1}du\right| \nonumber\\
             &\leq&
        \frac{\mu_{s_0}^{v_0}c_0s_0 |u_{s_0}|}{
          |\phi^{(s_0)}(\theta_0)|} \int_0^{1}
               u^{v_0}(  1-   u)^{s_0-1}d
              u
           \leq
          \frac{\mu_{s_0}^{v_0}c_0s_0 |u_{s_0}|}{
          |\phi^{(s_0)}(\theta_0)|}\frac{\Gamma(v_0+1){\Gamma(s_0)}}{\Gamma(s_0+1+v_0)}
          \leq
          \varepsilon.
\end{eqnarray*}
This completes the proof of Proposition \ref{Proposition:poly for nn
plus poly}. $\Box$

\subsection{Approximation of univariate polynomials by neural networks and the product gate}
 Our second tool, to be presented in the following proposition, shows that the approximation capability of shallow nets is not worse than that of polynomials of the same order (degree $+1$) as the cardinality of weights of the shallow nets.

\begin{proposition}\label{Proposition:polynomial for nn}
Under Assumption \ref{Assumption:smooth1} with $r_0=s_0+v_0$ and
$\theta_0\in\mathbb R$, let $k\in\{0,\dots,s_0\}$ and  $p_k(t)=\sum_{i=0}^ku_{i}t^i$. Then for an arbitrary
$\varepsilon\in(0,1)$, there exists a shallow net
$$
       h_{k+1}(t):=\sum_{j=1}^{k+1}a_j\phi(w_j\cdot t +\theta_0)
$$
with $0< w_j\leq 1$ and
\begin{equation}\label{bound of weights poly}
     |a_j|\leq
     \tilde{C}_1\left\{\begin{array}{cc}
     \left(1+\sum_{i=0}^{k}|u_{i}|\right)^{(k+1)!}\varepsilon^{-(k+1)!},
     &  \mbox{if}\quad 0\leq k\leq s_0-1,\\
     \left(1+\sum_{i=0}^{s_0}|u_{i}|\right)^{(1+s_0/v_0)s_0!}\varepsilon^{-(1+s_0/v_0)s_0!},
     &  \mbox{if}\quad k=s_0,
     \end{array}
     \right.
\end{equation}
for $1\leq j\leq k+1$, such that
\begin{equation}\label{error estimate pk h}
       |p_k(t)-h_{k+1}(t)|\leq  \varepsilon, \qquad
        \forall\ t\in [-1,1],
\end{equation}
where $\tilde{C}_1\geq1$ is a constant depending only on $\phi$,
$\theta_0$, $v_0$ and $s_0$,  to be specified explicitly in the
proof of the derivation.
\end{proposition}

We remark, however, that to arrive at a fair comparison with
polynomial approximation, the polynomial degree $k$ should be
sufficiently large, so that the norm of  weights of the shallow
nets could also be extremely large. In the following discussion, we
require $k$ to be independent of $\varepsilon$ in order to reduce
the norm of the weights. Based on Proposition
\ref{Proposition:polynomial for nn}, we are able to derive the
following proposition, which yields a ``product-gate''  property of
deep nets.

\begin{proposition}\label{Proposition:product gate}
Under Assumption \ref{Assumption:smooth1} with $r_0=s_0+v_0$ and
$\theta_0\in\mathbb R$, for $\varepsilon\in(0,1)$, there exists a
shallow net
$$
       h_{3}(t):=\sum_{j=1}^{3}a_j\phi(w_j\cdot t +\theta_0)
$$
with
\begin{equation}\label{bound of weight 2.22}
    0< w_j\leq 1,\qquad |a_j|\leq
      \tilde{C}_2\left\{\begin{array}{cc}
      \varepsilon^{-6},
     &  \mbox{if}\  s_0\geq 3,\\
      \varepsilon^{-\frac{ 6 }{v_0}},
     &  \mbox{if}\ s_0=2
     \end{array} \right.
\end{equation}
for $j=1,2,3$, such that for any $U,U'\in[-1,1]$,
\begin{equation}\label{product gate}
       |UU'-(2h_{3}((U+U')/2)-h_{3}(U)/2-h_{3}(U')/2)|\leq
       \varepsilon,
\end{equation}
where $\tilde{C}_2$ is a constant depending only on $s_0$, $v_0$,
$\phi$ and $\theta_0$.
\end{proposition}

{\bf Proof.} For $\varepsilon>0$, we apply Proposition \ref{Proposition:polynomial for nn} to the polynomial $t^2$ to derive a shallow net
$$
        h_{3}(t)=\sum_{j=1}^{3}a_j\phi(w_j\cdot t +\theta_0)
$$
with $0< w_j\leq 1$ and
\begin{equation}\label{proof 1}
     |a_j|\leq
     \tilde{C}_1\left\{\begin{array}{cc}
      2^6\varepsilon^{-6},
     &  \mbox{if}\  s_0\geq 3,\\
      2^{\frac6{v_0}}\varepsilon^{-\frac{ 6 }{v_0}},
     &  \mbox{if}\ s_0=2
     \end{array}
     \right.
\end{equation}
for $j=1,2,3$, such that
\begin{equation}\label{proof 2}
     |t^2-h_{3}(t)|\leq \varepsilon,\qquad t\in[-1,1].
\end{equation}
Since
$$
      UU'=\frac{4\left(\frac{U+U'}2\right)^2-U^2-(U')^2}{2}
$$
and $U,U'\in[-1,1]$ implies $(U+U')/2\in[-1,1]$,  we have
$$
       |h_{3}((U+U')/2)-((U+U')/2)^2|\leq \varepsilon,\qquad
       |h_{3}(U)-U^2|\leq \varepsilon,\qquad |h_{3}(U')-(U')^2|\leq
       \varepsilon.
$$
This completes the proof of Proposition \ref{Proposition:product
gate} by scaling $\varepsilon$ to $\varepsilon/3$. $\Box$

To end this subsection, we   present the proof of Proposition
\ref{Proposition:polynomial for nn}.

{\bf Proof of Proposition \ref{Proposition:polynomial for nn}.}
Observe that $\frac{1}{\min\{1,a\}}=\max\left\{1,\frac1a\right\}$
for $a>0$ and
$\max\left\{1,\frac{|u_k|}\varepsilon\right\}\leq\max\left\{1,\left(\frac{|u_k|}\varepsilon\right)^{1/v_0}\right\}$.
For the case $k=s_0$, the constant $\mu_k=\mu_{k,\varepsilon}$
defined by (\ref{muk}) satisfies
$$
   \frac1\mu_k\leq C_{\phi,s_0}\max\left\{1,\left(\frac{|u_k|}\varepsilon\right)^{1/v_0}\right\},
$$
where $C_{\phi,s_0}$ is a constant depending on $\phi$ and $s_0$ and
given by
$$
       C_{\phi,s_0}=\max\left\{\max_{1\leq k\leq s_0-1}\frac{\|\phi^{(k+1)}\|_{C[\theta_0-1,\theta_0+1]}}{|\phi^{(k)}(\theta_0)|(k+1)},
       \left(\frac{s_0!\Gamma(v_0+1)c_0}{|\phi^{(s_0)}(\theta_0)|\Gamma(s_0+v_0+1)}\right)^{1/v_0}\right\}.
$$
For $0\leq i\leq k-1$,   the $i$-th coefficient of the
polynomial $p_{k-1}^*$ is bounded by
\begin{eqnarray*}
    &&
    |u_i|+\frac{|u_k|k!|\phi^{(i)}(\theta_0)|}{|\phi^{(k)}(\theta_0)|i!}C_{\phi,s_0}^{k-i}\max\left\{
    1,\left(\frac{|u_k|}\varepsilon\right)^{\frac{k-i}{v_0}}\right\}\\
    &\leq&
    \left(1+\frac{\sum_{i=0}^{k-1}|\phi^{(i)}(\theta_0)|}{|\phi^{(k)}(\theta_0)|}k!\right)(1+C_{\phi,s_0})^k
    \|u\|_1\max\left\{
    1,\left(\frac{\|u\|_1}\varepsilon\right)^{\frac{k}{v_0}}\right\}\\
    &\leq&
    \tilde{C}_k\|u\|_1\max\left\{
    1,\left(\frac{\|u\|_1}\varepsilon\right)^{\frac{k}{v_0}}\right\},
\end{eqnarray*}
where $\|u\|_1=\sum_{i=0}^k|u_i|$  and the constant $\tilde{C}_k$ is
given by
$$
     \tilde{C_k}= \left(1+\frac{\sum_{i=0}^{k-1}|\phi^{(i)}(\theta_0)|+1}{|\phi^{(k)}(\theta_0)|}k!\right)
     (1+C_{\phi,s_0})^k.
$$
{  Also,} the coefficient of $\phi(\mu_kt+\theta_0)$ in
(\ref{error estimate for p bb n p}) satisfies
$$
      \left|u_k\frac{k!}{\mu_k^k\phi^{(k)}(\theta_0)}\right|\leq\tilde{C}_k\|u\|_1\max\left\{
    1,\left(\frac{\|u\|_1}\varepsilon\right)^{\frac{k}{v_0}}\right\}.
$$
Denote $C'_{s_0}=\max_{0\leq k\leq s_0}\tilde{C}_k(k+1)^{k/v_0}.$
Then it follows from Proposition \ref{Proposition:poly for nn plus
poly}, with $\varepsilon$ scaled to $\frac{\varepsilon}{k+1}$, that
$$
    \max_{-1\leq t\leq 1}|p_k(t)-a_1\phi(w_1t+\theta_0)-p_{k-1}^*(t)|\leq\frac{\varepsilon}{k+1},
$$
where $p_{k-1}^*(t)=\sum_{i=0}^{k-1}c_it_i$ satisfies $|c_i|\leq
C'_{s_0}\|u\|_1^{\frac{k}{v_0}+1}\varepsilon^{-\frac{k}{v_0}}$ for
$i=0,\dots,k-1$,$w_1\in (0,1]$ and $|a_1|\leq
C'_{s_0}\|u\|_1^{\frac{k}{v_0}+1}\varepsilon^{-\frac{k}{v_0}}.$ If
the leading term of $p_{k-1}^*(t)$ is $c_{i_0}t^{i_0}$ with $0\leq
i_0\leq k-1$, then we may apply Proposition \ref{Proposition:poly
for nn plus poly} with $\frac{\varepsilon}{k+1}$ and $v_0=1$ again
to obtain
$$
    \max_{-1\leq t\leq 1}|p_{k-1}^*(t)-a_2\phi(w_2t+\theta_0)-p^*_{i_0-1}(t)|\leq\frac{\varepsilon}{k+1},
$$
where  $w_2\in(0,1]$, and $a_2$  as well as the coefficient $c_i^*$
of $p_{i_0-1}^*(t)=\sum_{i=0}^{i_0-1}c_i^*t^i$  are bounded above by
$$
     C_{s_0}'\left(kC_{s_0}'\|u\|_1^{\frac{k}{v_0}+1}\varepsilon^{-\frac{k}{v_0}}\right)^{i_0+1}\varepsilon^{-i_0}
     \leq k^k(C_{s_0}')^{1+k}\|u\|_1^{k\left(\frac{k}{v_0}+1\right)}\varepsilon^{-\frac{k^2}{v_0}-k+1}.
$$
Then our conclusion follows by mathematical induction with the
constant $\tilde{C}_1$ given by
$\tilde{C_1}=k^{k+1}(C_{s_0}')^{k^k}.$ The case $k\leq s_0-1$ can be
easily verified {  with the same procedure}. This completes the proof of Proposition
\ref{Proposition:polynomial for nn}.
$\Box$

\subsection{Approximating smooth functions by products of polynomials and neural networks}
In this subsection, we discuss the approximation of continuous
functions on $\mathbb J:=[0,1/2]$ by sums of the products of Taylor
polynomials and shallow nets. Let $n\in\mathbb N$ and
$t_j=\frac{j}{2n}$ with $j=0,1,\dots,n$  be the equally spaced
points on $\mathbb J$. For an arbitrary $t\in\mathbb J,$ there is
some $j_0,$ such that $t_{j_0}\leq t<t_{j_0+1}$ ($t_{n-1}\leq t\leq
t_n$ when $t=1/2$).  Since
$$
       -4An(t-t_j)+A\leq -A \qquad \mbox{for}\quad
       j=0,1,\dots,j_0-1,
$$
and
$$
       -4An(t-t_j)+A> A \qquad \mbox{for}\quad
       j=j_0+1,j_0+2,\dots,n,
$$
we  may derive from (\ref{def delta111})  the following localized
approximation property:
\begin{equation}\label{Localized approximation}
    \left\{\begin{array}{cc}
    |\phi(-4An(t-t_j)+A)|\leq \delta_\phi(A),& \mbox{if}\ j\leq
    j_0-1,\\
     |\phi(-4An(t-t_j)+A)-1|\leq \delta_\phi(A), &\ \ \ \ \ \mbox{if}\
    j_0+1\leq j\leq n.\end{array}\right.
\end{equation}
For a purpose of approximation theory, we need the following  error estimate of
the Taylor expansion  which is an easy consequence of Lemma
\ref{Lemma:Taylor formula}.

\begin{lemma}\label{Lemma:Taylor approximation}
Let $\psi\in Lip^{(r,c_0')}_{\mathbb J}$  with $r=s+v$, $s\in\mathbb
N_0$, $0<v\leq1$ and $c_0'>0$. Define
$$
       T_{s,\psi,\tilde{t}}(t):=\sum_{j=0}^{s}\frac{\psi^{(j)}(\tilde{t})}{j!}(t-\tilde{t})^j.
$$
Then
\begin{equation}\label{Taylor approximation}
     |\psi(t)-T_{s,\psi,\tilde{t}}(t)|\leq \frac{c_0'}{s!}|t-\tilde{t}|^r,\qquad\forall\ t,\tilde{t}\in \mathbb J.
\end{equation}
\end{lemma}

With the localized approximation property (\ref{Localized
approximation}) and Lemma \ref{Lemma:Taylor approximation}, for each
$g\in Lip_{\mathbb J}^{(r,c_0')}$, we now define
\begin{equation}\label{operator2}
                   \Phi_{n,s,g,A}(t):=\sum_{j=0}^nT_{s,g,t_j}(t)b_{A,j}(t),
\end{equation}
where
$$
           b_{A,0}(t):=\phi\left(-4Ant+A\right),
$$
and
$$
            b_{A,j}(t):=\phi\left(-4An(t-t_j)+A\right)-\phi\left(-4An(t-t_{j-1})
            +A\right),\qquad1\leq j\leq n.
$$
Note that each term in the approximant (\ref{operator2}) is the
product of a Taylor polynomial and a shallow neural network
function, with the special case of  $s=0$ already considered in
\cite{Chen1993}. We provide an error estimate for $\Phi_{n,s,g,A}$
in the following proposition.

\begin{proposition}\label{Proposition:jackson1}
If $g\in Lip_{\mathbb J}^{(r,c_0')}$ {  with $r=s+v, s\in\mathbb N_0, 0<v\leq 1, c_0'>0$} and $\phi$ is a bounded
sigmoidal function, then
$$
       |g(t)-\Phi_{n,s,g,A}(t)|\leq  \tilde{C}_{3}(n\delta_\phi(A)
            +n^{-r}),\qquad\forall\ t\in\mathbb J,
$$
where
$\tilde{C}_{3}:=2\left(\frac{c_0'+c_0'\|\phi\|_{L_\infty(\mathbb
R)}}{s!}+\|g\|_{L_\infty(\mathbb J)}\right)$.
\end{proposition}

{\bf Proof.} For  $t\in\mathbb J$, let $j_0$ be the integer that satisfies $t_{j_0}\leq t<t_{j_0+1}$ for $0\leq j_0\leq n-2,$ and $t_{j_0}\leq t\leq t_{j_0+1}$ for $j_0=n-1,$ while ${  t_{n-1}}\leq t\leq t_n$ if $t=1/2$. Then by separating $\sum_{j=0}^n$ into $\sum_{j=0}^{j_0}+\sum_{j_0+1}^n$, it follows
from (\ref{operator2}) that
\begin{eqnarray*}
   \Phi_{n,s,g,A}(t)&=&\sum_{j=0}^{j_0}(T_{s,g,t_j}(t)-T_{s,g,t_{j+1}}(t))
          \phi\left(-4An(t-t_j)+A\right)\\
          &+&
         \sum_{j=j_0+1}^{n-1}(T_{s,g,t_j}(t)-T_{s,g,t_{j+1}}(t))
          \left(\phi\left(-4An(t-t_j)+A\right)-1\right)\\
          &+&
          T_{s,g,t_n}(t)\left(\phi\left(-4An(t-t_n)+A\right)-1\right)
          +T_{s,g,t_{j_0+1}}(t),
\end{eqnarray*}
where the last term appears because the term
$T_{s,g,t_{j_0+1}}(t)b_{A,j_0+1}(t)$  is separated in
(\ref{operator2}) into the above summations. It follows by
considering the term with $j=j_0$ from the first summation that
\begin{eqnarray*}
           \left|g(t)-\Phi_{n,s,g,A}(t)\right|
           &\leq&
           \sum_{j=0}^{j_0-1}|T_{s,g,t_j}(t)-T_{s,g,t_{j+1}}(t)|
          \left|\phi\left(-4An(t-t_j)+A\right)\right|\\
          &+&
          \sum_{j=j_0+1}^{n-1}|T_{s,g,t_j}(t)-T_{s,g,t_{j+1}}(t)|
          \left|\phi\left(-4An(t-t_j)+A\right)-1\right|\\
            &+&
                  |T_{s,g,t_n}(t)|\left|\phi\left(-4An(t-t_n)+A\right)-1\right|
                  +|T_{s,g,t_{j_0+1}}(t)-g(t)|\\
                  &+&
           |T_{s,g,t_{j_0}}(t)-g(t)+g(t)-T_{s,g,t_{j_0+1}}(t)|
          \left|\phi\left(-4An(t-t_{j_0})+A\right)\right|.
\end{eqnarray*}
Noting (\ref{Localized approximation}) and Lemma \ref{Lemma:Taylor
approximation}, we have
\begin{eqnarray*}
           \left|g(t)-\Phi_{n,s,g,A}(t)\right|
           &\leq&
           (2n-1) \max_{0\leq j\leq
          n}|T_{s,g,t_j}(t)|\delta_\phi(A)+\frac{c_0'}{s!}(1+2\|\phi\|_{L_\infty(\mathbb
          R)})n^{-r}.
\end{eqnarray*}
On the other hand, since  (\ref{Taylor
approximation}) implies
$$
  \max_{0\leq t\leq 1,0\leq j\leq
      n}|T_{s,g,t_j}(t)|\leq \frac{c_0'}{s!}+\|g\|_{L_\infty(\mathbb J)},
$$
we have
\begin{eqnarray*}
           \left|g(t)-\Phi_{n,s,g,A}(t)\right|
           &\leq&
           (2n-1)\left(\frac{c_0'}{s!}+\|g\|_{L_\infty(\mathbb
           J)}\right)
           \delta_\phi(A)+\frac{c_0'}{s!}(1+2\|\phi\|_{L_\infty(\mathbb
           R)})n^{-r}.
\end{eqnarray*}
This completes the proof of Proposition \ref{Proposition:jackson1}.
$\Box$

\subsection{Approximation of univariate functions by neural networks with two hidden layers}

Based on  Propositions \ref{Proposition:polynomial for nn},
\ref{Proposition:product gate} and \ref{Proposition:jackson1}, we
prove the following theorem on the construction of deep nets with
two hidden layers for the approximation of univariate smooth
functions.

\begin{theorem}\label{Theorem:app phi n s A}
Let $g\in Lip_{\mathbb J}^{(r,c_0')}$ with { $c_0'>0$, $r=s+v$, $s\in\mathbb N_0, 0<v\leq 1$. Then under
Assumption \ref{Assumption:smooth1} with $c_0>0$, $r_0=s_0+v_0$, $0<v_0\leq 1$,} and
$s_0\geq \max\{s,2\}$, for an arbitrary $0<\varepsilon\leq 1$, there
exists a deep net of the form
\begin{equation}\label{Def H 2 hidden}
       H_{3(n+3),s+3,A}(t)=
       \sum_{j=1}^{3n}a^*_j\phi\left(\sum_{i=1}^{s+3}a^*_{j,i}\phi(w^*_{j,i}t+\theta^*_{j,i})+\theta_j^*\right),\qquad
       t\in\mathbb J
\end{equation}
that satisfies $|\theta_j^*|, |\theta^*_{j,i}|\leq
1+3An+|\theta_0|$, $|w_{j,i}^*|\leq 4An$ and
\begin{equation}\label{bound weight deep1}
      |a_j^*|,|a^*_{j,i}|\leq \tilde{C}_{4}\left\{\begin{array}{cc}
      \varepsilon^{-7(s+1)!},
     &  \mbox{if}\  s_0\geq 3, s_0> s\\
       \varepsilon^{-\frac{7 }{v_0}(s+1)!},
     &  \mbox{if}\ s_0\geq 3,s_0=s,\\
      \varepsilon^{-\frac{v_0+6}{v_0} (s+1)!},
     &  \mbox{if}\  s_0=2, s_0>s,\\
       \varepsilon^{-\frac{v_0+6 }{v_0^2}(s+1)!},
     &  \mbox{if}\ s_0=2, s_0=s
     \end{array}\right.
\end{equation}
such that
\begin{equation}\label{Jackson for 2}
       |g(t)-H_{3n+3,s+3,A}(t)|
       \leq
        \tilde{C}_{4}(n\delta_\phi(A)
            +n^{-r}+n\varepsilon),\qquad\forall \ t\in\mathbb J,
\end{equation}
for some constant $\tilde{C}_{4}$ independent of $\varepsilon$, $n$
and $A$.
\end{theorem}

The main novelty of the above theorem is that (\ref{Jackson for 2})
holds for an arbitrary $r>0$ and the parameters of the deep net
(\ref{Def H 2 hidden}) are controllable, provided that the
activation function satisfies Assumption \ref{Assumption:smooth1}
with $r_0\ge r$. This deviates Theorem \ref{Theorem:app phi n s A}
from the classical results in \cite{Chen1993, Chui1994, Mhaskar1996,
Xie2010,Ying2017}, in which either $0<r\leq 1$ is required or extremely large
parameters are needed. We remark that since the goal of this paper
is to  approximate radial functions, we only need error estimates
for approximation of univariate functions, {  though} the approach in
this paper can be extended to the realization of more general
multivariate functions by standard arguments.

{\bf .Proof of Theorem \ref{Theorem:app phi n s A}} The proof of this theorem is divided into three steps: first to decouple the product, then to approximate the Taylor polynomials, and finally to deduce the approximation errors, by applying Propositions \ref{Proposition:product gate}, \ref{Proposition:polynomial for nn}, and \ref{Proposition:jackson1}, respectively.

{\it Step 1: Decoupling  products.} From Assumption
\ref{Assumption:smooth1}, the definition of $b_{A,j}$, and Lemma
\ref{Lemma:Taylor approximation}, we observe that
$$
    |b_{A,j}(t)|\leq 2, \qquad
    |T_{s,g,t_j}(t)|\leq \|g\|_{L_\infty(\mathbb J)}+ c_0',\qquad\forall\ t,t_j\in \mathbb J.
$$
By denoting
$$
    B_1:=4(\|g\|_{L_\infty(\mathbb J)}+c_0'+2)
$$
we have, for an arbitrary $t\in\mathbb J$,
$b_{A,j}(t)/B_1,T_{s,g,t_j}(t)/B_1\in[-1/4,1/4]$. It then follows
from Proposition \ref{Proposition:product gate} with
$U=b_{A,j}(t)/B_1$ and $U'=T_{s,g,t_j}(t)/B_1$ that a shallow net
$$
       h_{3}(t):=\sum_{j=1}^{3}a_j\phi(w_j\cdot t +\theta_0)
$$
can be {  constructed} to satisfy the conditions $0< w_j\leq 1$ and the
bound (\ref{bound of weight 2.22}) for $a_j$ that depends only on
$\varepsilon$, such that
\begin{equation}\label{first argument}
       \left|T_{s,g,t_j}(t)b_{A,j}(t)-B_1^2\left(2h_{3}\left(\frac{T_{s,g,t_j}(t)+b_{A,j(t)}}{2B_1}\right)
       -\frac{h_{3}\left(\frac{b_{A,j}(t)}{B_1}\right)}2-\frac{h_{3}\left(\frac{T_{s,g,t_j}(t)}{B_1}\right)}2
       \right)\right|
       \leq
       B_1^2\varepsilon.
\end{equation}
Furthermore, it follows from (\ref{bound of weight 2.22}) and
$\|\phi'\|_{L^\infty(\mathbb R)}\leq 1$ that for any
$\tau,\tau'\in\mathbb J$,
\begin{equation}\label{lip1}
    | h_3(\tau)-h_3(\tau')|\leq
    \sum_{j=1}^3|a_j||\tau-\tau'|\leq
    3\tilde{C}_{2} |\tau-\tau'|
      \left\{\begin{array}{cc}
      \varepsilon^{-6},
     &  \mbox{if}\  s_0\geq 3,\\
      \varepsilon^{-\frac{ 6 }{v_0}},
     &  \mbox{if}\ s_0=2.
     \end{array}
     \right.
\end{equation}

{\it Step 2: Approximating Taylor  polynomials.} Since
$t,t_j\in\mathbb J$, we have $t-t_j\in[-1,1]$. Let
$\varepsilon_1\in(0,1/4]$ to be determined later. Then, for {  any}
fixed $j\in\{1,2,\dots,n\}$, it follows from Proposition
\ref{Proposition:polynomial for nn}  with $p_s(t-t_j) =
T_{s,g,t_j}(t)/B_1=\sum_{i=0}^{s}\frac{g^{(i)}(t_j)}{i!B_1}(t-t_j)^i$
that there exists a shallow net
\begin{equation}\label{Def hs+1}
       h_{s+1,j}(t):=\sum_{i=1}^{s+1}a_{i,j}\phi(w_{i,j}\cdot t-w_{i,j}t_j+\theta_0 )
\end{equation}
with $0< w_{i,j}\leq 1$ and
\begin{equation}\label{bound of weights poly Taylor}
     |a_{i,j}|\leq
     \tilde{C}_{5}\left\{\begin{array}{cc}
      \varepsilon_1^{-(s+1)!},
     &  \mbox{if}\  s_0>s,\\
      \varepsilon_1^{-(s_0/v_0+1)s_0!},
     &  \mbox{if}\ s_0=s,
     \end{array}
     \right.
\end{equation}
where $\tilde{C}_{5}:=\tilde{C_1}\left(1+\sum_{i=0}^{s}
\left(\frac{\|g^{(i)}\|_{L^\infty(\mathbb
J)}}{i!B_1}\right)\right)^{(s_0/v_0+1)s_0!}$, such that
\begin{equation}\label{induction 5555}
       |T_{s,g,t_j}(t)/B_1-h_{s+1,j}(t)|\leq  \varepsilon_1, \qquad
       1\leq j\leq n,\  \forall\ t\in \mathbb J.
\end{equation}

{\it Step 3: Construction of deep nets with error bounds.} Define
\begin{equation}\label{Def H}
   H_{3n+3,s+3,A}(t):=\sum_{j=0}^nH_{A,j}(t)
\end{equation}
with
\begin{equation}\label{Def Hj}
     H_{A,j}(t):= B_1^2 \left[2h_3\left(\frac{h_{s+1,j}(t)}{2}+\frac{b_{A,j}(t)}{2B_1}\right)
     -\frac{h_3\left(h_{s+1,j}(t)\right)}2-\frac{h_3\left(\frac{b_{A,j}(t)}{B_1}\right)}2\right].
\end{equation}
{  Then it follows from (\ref{operator2}) and (\ref{first argument})
that
\begin{eqnarray}\label{s.1}
  &&\left|H_{3n+3,s+3,A}(t)-\Phi_{n,s,g,A}(t)\right|
  \leq
  \sum_{j=0}^n\left|H_{A,j}(t)-T_{s,g,t_j}(t)b_{A,j}(t)\right| \nonumber\\
  &\leq&
  \sum_{j=0}^n\left|H_{A,j}(t)-B_1^2\left(2h_{3}\left(\frac{T_{s,g,t_j}(t)+b_{A,j}(t)}{2B_1}\right)
       -\frac{h_{3}\left(\frac{b_{A,j}(t)}{B_1}\right)}2-\frac{h_{3}\left(\frac{T_{s,g,t_j}(t)}{B_1}\right)}2
       \right)\right| \nonumber\\
       &+&
       (n+1)B_1^2\varepsilon.
\end{eqnarray}
}
Also, since $0<\varepsilon_1\leq1/4$ and $T_{s,g,t_j}(t)/B_1\leq
1/4$, it follows from (\ref{induction 5555}) and (\ref{lip1}) that
\begin{eqnarray*}
   &&
   \left|h_3(h_{s+1,j}(t))-h_3(T_{s,g,t_j}(t)/B_1)\right|
   \leq
   3\tilde{C}_{2} \varepsilon_1
      \left\{\begin{array}{cc}
      \varepsilon^{-6},
     &  \mbox{if}\  s_0\geq 3,\\
      \varepsilon^{-\frac{ 6 }{v_0}},
     &  \mbox{if}\ s_0=2,
     \end{array}
     \right.
\end{eqnarray*}
and
\begin{equation*}
  \left|h_3\left(\frac{h_{s+1,j}(t)}{2}+\frac{b_{A,j}(t)}{2B_1}\right)
  -h_{3}\left(\frac{T_{s,g,t_j}(t)+b_{A,j}(t)}{2B_1}\right)\right|
  \leq
  \frac{3\tilde{C}_{2}}2 \varepsilon_1
      \left\{\begin{array}{cc}
      \varepsilon^{-6},
     &  \mbox{if}\  s_0\geq 3,\\
      \varepsilon^{-\frac{ 6 }{v_0}},
     &  \mbox{if}\ s_0=2.
     \end{array}
     \right.
\end{equation*}
Therefore, plugging the above two estimates into (\ref{s.1}), we
obtain for any $t\in {\mathbb J}$
\begin{equation}\label{s.2}
   \left|H_{3n+3,s+3,A}(t)-\Phi_{n,s,g,A}(t)\right|
   \leq
   (n+1)B_1^2\varepsilon+\frac{9\tilde{C}_{2}B_1^2}2 \varepsilon_1
      \left\{\begin{array}{cc}
      \varepsilon^{-6},
     &  \mbox{if}\  s_0\geq 3,\\
      \varepsilon^{-\frac{ 6 }{v_0}},
     &  \mbox{if}\ s_0=2.
     \end{array}
     \right.
\end{equation}

From the above argument,  we may set
$\varepsilon_1=\left\{\begin{array}{cc}
      \frac14\varepsilon^{7},
     &  \mbox{if}\  s_0\geq 3,\\
      \frac14\varepsilon^{1+\frac{ 6 }{v_0}},
     &  \mbox{if}\ s_0=2\end{array}\right.$
so that  (\ref{s.2}) implies that for any $t\in {\mathbb J}$
$$
   \left|H_{3n,s+3,A}(t)-\Phi_{n,s,g,A}(t)\right|
   \leq
   (n+1)(B_1^2+\frac98\tilde{C}_{2}B_1^2)\varepsilon.
$$
Applying this together with Proposition \ref{Proposition:jackson1},
we may conclude, for any  $t\in\mathbb J$, that
\begin{eqnarray*}
       &&|g(t)-H_{3n,s+3,A}(t)|\leq
       |g(t)-\Phi_{n,s,g,A}(t)|+|\Phi_{n,s,g,A}(t)-H_{3n,s+3,A}(t)|\\
       &\leq&
        (\tilde{C}_{3}+B_1^2+9\tilde{C}_{2}B_1^2/8)((n+1)\delta_\phi(A)
            +n^{-r}+(n+1)\varepsilon).
\end{eqnarray*}
What is left is to find   bounds of the parameters in
$H_{3n,s+3,A}$. This can be done by applying (\ref{Def H}),
(\ref{Def Hj}), (\ref{Def hs+1}), the definition of $b_{A,j}$,
(\ref{bound of weight 2.22}) and (\ref{bound of weights poly
Taylor}) to yield
$$
       H_{3n+3,s+3,A}(t)=\sum_{j=1}^{3n+3}a^*_j\phi\left(\sum_{i=1}^{s+3}a^*_{j,i}\phi(w^*_{j,i}t+\theta^*_{j,i})+\theta_j^*\right),
$$
by considering $|\theta_j^*|, |\theta^*_{j,i}|\leq
1+3An+|\theta_0|$, $|w_{j,i}^*|\leq 4An$, with $a_j^*,a^*_{j,i}$ to
satisfy (\ref{bound weight deep1}) for the constant
$\tilde{C}_{4}:=\tilde{C}_{3}+B_1^2+9\tilde{C}_{2}B_1^2/8+1$, which
is independent of $\varepsilon$, $n$ or $A$. This completes the
proof of Theorem  \ref{Theorem:app phi n s A}.
$\Box$

\section{Proofs of Main Results}\label{Sec.Proof}
This section is devoted to proving our main results, to be presented
in three subsections, namely:  Proof of Theorem \ref{Theorem:lower
bound for deep nets 1}, Proofs of Theorem \ref{Theorem:almost
optimal}, and Proof of Theorem \ref{Theorem: ERM}, respectively.

\subsection{Proof of Theorem \ref{Theorem:lower bound for deep nets 1}}
Our proof of Theorem \ref{Theorem:lower bound for deep nets 1} will
require two mathematical {  tools on  relationships} among covering
numbers \cite{Zhou2002,Zhou2003}, lower bounds of  approximation,
and an upper bound estimate for the covering number of $\mathcal
H^{tree}_{L,\alpha,\mathcal R}$. It is well-known that the
approximation capability of a class of functions depends on its
``capacity" (see, for example, \cite{Maiorov1999c}). In the
following lemma, we will establish some relationship between
covering numbers and lower bound of approximation, when the target
function is radial.

\begin{lemma}\label{Lemma:Relation c and l}
Let $N\in\mathbb N$ and $V\subseteq L_\infty(\mathbb B^d)$. If
{  \begin{equation}\label{covering condition}
  \mathcal N(\varepsilon,V)\leq
  C_1'\left(\frac{C_2'N^\beta}{\varepsilon}\right)^N,\qquad \forall   0<\varepsilon\leq 1
\end{equation}}
with $\beta,C_1',C_2'>0$, then
\begin{equation}\label{lower bound deep}
   \mbox{dist}(Lip^{(\diamond,r,c_0)},V,L_\infty(\mathbb B^d))\geq
   C_3'\left[N\log_2(N+C_4')\right]^{-r},
\end{equation}
where $\mathcal N(\varepsilon,V)$ denotes the $\varepsilon$-covering
number of $V$ {  in $L_\infty(\mathbb B^d)$}, which is the least number of elements in an
$\varepsilon$-net of $V$, $C_3'=\frac{c_0}8(\beta+2r+4)^{-r}$ and
$C_4'=2C_1'+4C_2'c_0^{-1}(\beta+2r+4)^r$.
\end{lemma}
The proof of Lemma \ref{Lemma:Relation c and l} is motivated by
\cite{Maiorov1999c}, where a relation between the pseudo-dimension
and lower bounds of approximating smooth functions was established.
We postpone its proof to Section \ref{Sec. Auxiliary Lemmas}. { The second relationship is } a tight bound {  for} covering numbers \cite{Chui2019}.

\begin{lemma}\label{Lemma:covering number}
Let $L\in\mathbb N$, $c_1>0$, and assume that $\phi_j\in
Lip^{(1,c_1)}_{\mathbb R}$ {  to satisfy $\|\phi_j\|_{L_\infty(\mathbb B^d)}\leq 1$, for
$j=0,\dots,L$. Then for any  $0<\varepsilon\leq 1$,}
\begin{equation}\label{covering number estimate:dn}
    \mathcal N(\varepsilon,\mathcal H^{tree}_{L,\alpha,\mathcal R})\leq
   \left(\frac{2^{ L+5/2}c_1^{ L+3/2}\mathcal
    A^{ L+1}_{R,\alpha,L}}{\varepsilon}\right)^{2\mathcal A_L},
\end{equation}
where $\mathcal A_{R,\alpha,L}:=\mathcal R \left(\mathcal
A_L\right)^\alpha$ and $\mathcal A_L$ is defined by (\ref{Def:AL}).
\end{lemma}

We are now ready
to prove Theorem \ref{Theorem:lower bound for deep nets 1} by
applying the above two lemmas.

{\bf Proof of Theorem \ref{Theorem:lower bound for deep nets 1}.}
In view of  Lemma \ref{Lemma:covering number}, {  condition
(\ref{covering condition}) is satisfied by $V=\mathcal
H^{tree}_{L,\alpha,\mathcal R}$ with} $\tilde{C}_1=1$, $N=2\mathcal A_L$,
$\beta=\alpha (L+1)$, and $\tilde{C}_2= 2^{L+5/2}c_1^{
L+3/2}\mathcal R^{L+1}$. Then it follows from (\ref{lower bound
deep}) that
\begin{eqnarray*}
   &&\mbox{dist}(Lip^{(\diamond,r,c_0)},\mathcal
     H^{tree}_{L,\alpha,\mathcal R},L_\infty(\mathbb B^d))\geq
   \frac{c_0}8( \alpha L+ \alpha+2r+4)^{-r}\\
   &\times&
   \left[2\mathcal A_L\log_2(2\mathcal A_L+2+2^{ L+9/2}c_0^{-1}c_1^{ L+3/2}\mathcal R^{ L+1}( \alpha L + \alpha+2r+4)^r)\right]^{-r}.
\end{eqnarray*}

Noting that
\begin{eqnarray*}
   &&\log_2(2\mathcal A_L+2+2^{ L+9/2}c_0^{-1}c_1^{ L+3/2}\mathcal
    R^{ L+1}( \alpha L + \alpha+2r+4)^r)\\
    &\leq&
    \log_2\left(4\mathcal
    A_L+8c_0^{-1}(2\alpha+2r+4)^r(2c_1\mathcal
    R)^{ L+3/2}L^r\right)\\
    &\leq&
    [1+\log_2(4+8c_0^{-1}(2\alpha+2r+4)^r)]\log_2(\mathcal A_L+(2c_1\mathcal
    R)^{ L+3/2}L^r),
\end{eqnarray*}
we may conclude that
\begin{equation}\label{ad1}
    \mbox{dist}(Lip^{(\diamond,r,c_0)},\mathcal
     H^{tree}_{L,\alpha,\mathcal R},L_\infty(\mathbb B^d))
     \geq
     \tilde{C}_1'\left[L\mathcal A_L\log_2\left(\mathcal A_L+(2c_1\mathcal
    R)^{ L+3/2}L^r\right)\right]^{-r},
\end{equation}
where
$$
  \tilde{ C}_1'=\frac{c_08^{-r}(\alpha+r+2)^{-r}}{8}[1+\log_2(4+8c_0^{-1}(2\alpha+2r+4)^r)]^{-r}.
$$
Next, {  for  $a>0$ and $\tilde{n}\geq 2$, it follows from direct computation
that
$$
   \log_2(\tilde{n}+a)\leq \log_2 [\tilde{n}(1+a)]\leq [1+\log_2(1+a)]\log_2 \tilde{n},
$$
which together  with  $a=(2c_1\mathcal R)^{L+3/2}L^r $ and
$\tilde{n}=\mathcal A_L$,  yields
\begin{eqnarray*}
  &&\log_2[\tilde{n}+ (2c_1\mathcal R)^{L+3/2}L^r] \leq  [1+\log_2( (2c_1\mathcal R)^{L+3/2}L^r+1)]\log_2\tilde{n}\\
  &\leq&  \log_2\tilde{n}+ (L+3/2)\log_2 (2c_1\mathcal R+1)\log_2\tilde{n}+r\log_2L\log_2\tilde{n}\\
    &\leq& [1+ \log_2 (2c_1\mathcal R+1)+r](L+3)\log_2\tilde{n}\\
   &\leq&
   4[1+r+\log_2 (2c_1\mathcal
  R+1)]L\log_2\tilde{n}.
\end{eqnarray*}
So, we have from (\ref{ad1}) that
\begin{equation}\label{lower 111}
   \mbox{dist}(Lip^{(\diamond,r,c_0)},\mathcal H^{tree}_{L,\alpha,\mathcal R},L_\infty(\mathbb B^d))\geq
   C_2^*(L^{2}\tilde{n}\log_2\tilde{n})^{-r}
\end{equation}
where $C_2^*:=\tilde{C}_1'[4[1+r+\log_2 (2c_1\mathcal R+1)]]^{-r}$.
This completes the proof of (\ref{lower bound for deep1}).

 The proof
(\ref{limitation for shallow}) is  easier. Let $\mathbb
S^{d-1}$ denote the unit sphere in $\mathbb R^d$, and consider the
manifold
$$
        \mathcal M_n:=\left\{\sum_{i=1}^na_i\phi_i(\xi_i\cdot x):\xi_i\in\mathbb S^{d-1}, \phi_i\in L^2([-1,1]),a_i\in\mathbb R\right\}
$$
of ridge functions. It is easy to see that $\mathcal
S_{\phi_1,n}\subset \mathcal M_n$. Then it follows from
\cite[Theorem 4]{Konovalov2008} there exist an integer
$\tilde{C}_2'$ and some positive real number $\tilde{C}_3'$, such that for
any   $f\in Lip^{(\diamond,r,c_0)}$},
$$
       \mbox{dist}(f,\mathcal S_{\phi_1,n},L_2(\mathbb B^d))\geq
        \mbox{dist}(f,\mathcal M_{n^{d-1}},L_2(\mathbb
        B^d))
        \geq
       \tilde{C}_3' \mbox{dist}(f,\mathcal P_{\tilde{C}_2'n}(\mathbb B^d),L_2(\mathbb
        B^d)),
$$
where $\mathcal P_s(\mathbb B^d)$ denotes the set of algebraic
polynomials defined on $\mathbb B^d$ of degrees not exceeding $s$.
But it was also proved in \cite[Theorem 1]{Konovalov2009} (with a
scaling of {  constants} in \cite[P.105]{Konovalov2009}), that
$$
      \mbox{dist}(Lip^{(\diamond,r,c_0)},\mathcal P_{\tilde{C}_2'n^{1/(d-1)}(\mathbb B^d))},L_2(\mathbb
      B^d))\geq \tilde{C}_4'n^{-r/(d-1)},
$$
where $\tilde{C}_4'$ is a constant depending only on $\tilde{C}_2'$,
$c_0$, $d$ and $r$. Therefore we have
$$
   \mbox{dist}(Lip^{(\diamond,r,c_0)},\mathcal S_{\phi_1,n},L_\infty(\mathbb B^d))
   \geq
   \mbox{dist}(Lip^{(\diamond,r,c_0)},\mathcal S_{\phi_1,n},L_2(\mathbb B^d))
   \geq
    C^*_1{\red \tilde{n}}^{-r/(d-1)}
$$
with {  $C^*_1:=\tilde{C}_3'\tilde{C}_4'/(d+2)$ by noting $\tilde{n}=(d+2)n$}. This establishes
(\ref{limitation for shallow}) and completes the proof of Theorem
\ref{Theorem:lower bound for deep nets 1}.
$\Box$

\subsection{Proof of Theorem \ref{Theorem:almost optimal}}

We shall show that based on Assumption \ref{Assumption:smooth1},
Theorem \ref{Theorem:almost optimal} is a consequence of the
following more general result, which we will first establish.

\begin{theorem}\label{Theorem:upper bound approximation}
Let $A\geq 1$. Under Assumption \ref{Assumption:smooth1} with
$r_0=s_0+v_0$, $s_0\geq 2$ and $0<v_0\leq 1$. Then for any $f\in
Lip^{(\diamond,r,c_0)}$ with $r\leq r_0$ {  and any $n\in\mathbb N$}, there is a deep net
\begin{eqnarray*}
       &&H_{3n+3,s+3,6,d,A}\\
       &=&
       \sum_{j=1}^{3n+3}a^*_j\phi\left(\sum_{i=1}^{s+3}a^*_{j,i}\phi\left(\sum_{k=1}^{6}a^*_{k,j,i}
       \phi\left(
       \sum_{\ell=1}^d
       a^*_{k,\ell,j,i}\phi\left(w^*_{k,\ell,j,i}  x^{(\ell)}+
       \theta^*_{k,\ell,j,i}\right)+\theta^*_{k,j,i}\right)
       +\theta^*_{j,i}\right)+\theta_j^*\right). \nonumber
\end{eqnarray*}
with $|w^*_{k,\ell,j,i}|\leq 1$,
{  $|\theta^*_{k,\ell,j,i}|,|\theta^*_{k,j,i}|,  |\theta^*_{j,i}|,|\theta_j^*|
     \leq 1+3An+|\theta_0|$
      and $|a_j^*|,|a^*_{j,i}|,|a^*_{k,j,i}|,|a^*_{k,\ell,j,i}|$
       bounded} by
$$
  \bar{C}_1\left\{\begin{array}{cc}
     (An^2)^{48} n^{48r(r+1)(1+7(s+1)!)},
     &  \mbox{if}\quad  s_0\geq 3, s_0> s\\
       (An^2)^{48} n^{48(r+1)(1+\frac{7 }{v_0})(s+1)!},
     &  \mbox{if}\quad s_0=s\geq 3,\\
      (An^2)^{\frac{ 6v_0+42 }{v_0}} n^{\frac{ (6v_0+42)(r+1) }{v_0}(1+\frac{v_0+6}{v_0} (s+1)!)},
     &  \mbox{if}\quad  s_0=2, s_0>s,\\
      (An^2)^{\frac{ 6v_0+42 }{v_0}} n^{\frac{ (6v_0+42)(r+1) }{v_0}(1+\frac{v_0+6}{v_0^2} (s+1)!)},
     &  \mbox{if}\quad s_0=s=2,
     \end{array}\right.
$$
such that{
\begin{equation}\label{Jaskson for deep nets}
       \|f-H_{3n,s+3,6,d,A}\|_{L_\infty(\mathbb B^d)}
       \leq \bar{C_2}(n\delta_\phi(A)+n^{-r}),
\end{equation}}
where $\delta_\phi(A)$ is defined by (\ref{def delta111}) and
$\bar{C}_1,\bar{C}_2$ are constants independent of $n$ or $A$.
\end{theorem}

{\bf Proof.}  We {  divide the proof} into four steps:
first to approximate $|{\bf x}|^2$,  next to unify the activation
function, then to construct the deep net, and finally to derive
bounds of the parameters.

{\it Step 1: Approximation of $|{\bf x}|^2$.} Since $f\in
Lip^{(\diamond,r,c_0)}$, there exists some $g^*\in
Lip^{(r,c_0)}_{\mathbb I}$  such that $f({\bf x})=g^*(|{\bf x}|^2)$.
Set $g(\cdot):=g^*(2\cdot)$. Then $f({\bf x})=g(|{\bf x}|^2/2)$ with
$g\in Lip_{\mathbb J}^{(2^rc_0,r)}$.  By Theorem \ref{Theorem:app
phi n s A},  for any  $0<\varepsilon\leq 1$, there is a deep net of
form (\ref{Def H 2 hidden})  such that
\begin{equation}\label{Jackson for 3}
       |f({\bf x}) - H_{3n+3,s+3,A}(|{\bf x}|^2/2)|
       \leq
        \tilde{C}_{4}(n\delta_\phi(A)
            +n^{-r}+n\varepsilon),\qquad\forall \ {\bf x}\in\mathbb
            B^d.
\end{equation}
{  We will first treat  components} $x^{(\ell)}$ of ${\bf x} =
(x^{(1)}, \cdots, x^{(d)})$ separately. {  Let
$0<\varepsilon_1\leq\frac1{d+2}$ to be determined below, depending on $\epsilon$. By
Proposition \ref{Proposition:polynomial for nn} applied to the quadratic polynomial $t^2$,
 there exists a shallow net
$$
       h_{3}(t):=\sum_{k=1}^{3}a_{k}\phi(w_{k}\cdot t+\theta_0 )
$$
with $0<w_{k}\leq 1$ and $|a_{k}|\leq
\tilde{C}_1\left\{\begin{array}{cc}
      2^6\varepsilon_1^{-6},
     &  \mbox{if}\  s_0\geq 3,\\
      2^{\frac6{v_0}}\varepsilon_1^{-\frac{ 6 }{v_0}},
     &  \mbox{if}\ s_0=2
     \end{array}
     \right.$
 such that
\begin{equation}\label{induction 66666}
       |t^2-h_{3}(t)|\leq  \varepsilon_1, \qquad
        \forall\ t\in \mathbb I.
\end{equation}
Hence, by setting
\begin{equation}\label{Def.h3d}
       h_{3d}({\bf x}):=\sum_{\ell=1}^dh_{3}(x^{(\ell)})/2
       =\sum_{k=1}^3\left(\sum_{\ell=1}^d\frac{a_{k}}2\phi(w_{k}\cdot x^{(\ell)}+\theta_0
       )\right),
\end{equation}
it follows from (\ref{induction 66666}) that
\begin{equation}\label{induction 77777}
       \left||{\bf x}|^2/2-h_{3d}({\bf x})\right|\leq  d\varepsilon_1/2, \qquad \forall\ {\bf x} \in \mathbb B^d.
\end{equation}
Hence, by the assumption $\|\phi\|_{L_\infty(\mathbb R)}\leq 1$, we have, for
${\bf x}\in\mathbb B^d$,
\begin{equation}\label{Def hua B}
      \left|\sum_{\ell=1}^d\frac{a_{k}}2\phi(w_{k}\cdot x^{(\ell)}+\theta_0
       )\right|\leq \frac12\sum_{\ell=1}^d|a_{k}|\leq
       { \tilde{C}_1}d\left\{\begin{array}{cc}
      2^6\varepsilon_1^{-6},
     &  \mbox{if}\  s_0\geq 3,\\
      2^{\frac6{v_0}}\varepsilon_1^{-\frac{ 6 }{v_0}},
     &  \mbox{if}\ s_0=2.
     \end{array}\right.
\end{equation}
}
In the following, we denote the above bound by $\mathcal B$ and note
that $\mathcal B \geq 1$.

{\it Step 2: Unifying the activation function.} From
(\ref{Def.h3d}), we note that $h_{3d}$ is a deep net with one hidden
layers. In this step, we will apply Proposition
\ref{Proposition:polynomial for nn} to unify the activation
functions. For any $\varepsilon_2\in(0,1)$ to be determined, it
follows from Proposition \ref{Proposition:polynomial for nn} {  applied to the linear function $t$}, with
$k=1$ and $s_0\geq 2$, that there exists a shallow net
$$
       h^*_{2}(t):=\sum_{k'=1}^{2}a_{k'}\phi(w_{k'}\cdot t +\theta_0)
$$
with
\begin{equation}\label{bound of weights inserting}
       0< w_{k'}\leq 1,\qquad \mbox{and}\quad
     |a_{k'}|\leq
     4\tilde{C}_1\varepsilon_2^{-6},
\end{equation}
 such that
\begin{equation}\label{error estimate pk h   inserting}
       |t-h_{2}^*(t)|\leq  \varepsilon_2, \qquad
        \forall\ t\in [-1,1].
\end{equation}
Inserting
{ $t=\frac{\sum_{\ell=1}^d\frac{a_{k}}2\phi(w_{k}\cdot
x^{(\ell)}+\theta_0)}{\mathcal B}$  into (\ref{error estimate pk h
inserting}), we have, for ${\bf x}\in\mathbb B^d$,
\begin{equation}\label{insert 1.1}
    \left|\sum_{\ell=1}^d\frac{a_{k}}2\phi(w_{k}\cdot x^{(\ell)}+\theta_0
       )-\mathcal Bh^*_{2}\left(\frac{\sum_{\ell=1}^da_{k}\phi(w_{k}\cdot x^{(\ell)}+\theta_0
       )}{2\mathcal B}\right)\right|
       \leq
       \mathcal B\varepsilon_2.
\end{equation}
Write
\begin{eqnarray}\label{Def. h6d}
       h_{6,d}(\bf x)
       &=&
       \sum_{k=1}^3\sum_{k'=1}^{2}\mathcal Ba_{k'}\phi\left(w_{k'}\frac{\sum_{\ell=1}^da_{k}\phi(w_{k}\cdot x^{(\ell)}+\theta_0
       )}{2\mathcal B}  +\theta_0\right) \nonumber\\
       &=:&
       \sum_{k=1}^6 a'_{k}\phi\left(\sum_{\ell=1}^da''_{k}\phi(w_{k}'\cdot x^{(\ell)}+\theta_0
       )  +\theta_0\right).
\end{eqnarray}}
It then follows from
 (\ref{bound of weights inserting}) and (\ref{Def hua B}) that {  $0<w_k'\leq 1$,
$$
     |a'_k|\leq d{ \tilde{C}_1}^2\varepsilon_2^{-6}\left\{\begin{array}{cc}
     2^6 \varepsilon_1^{-6},
     &  \mbox{if}\  s_0\geq 3,\\
     2^\frac{6}{v_0} \varepsilon_1^{-\frac{ 6 }{v_0}},
     &  \mbox{if}\ s_0=2,
     \end{array}\right.\qquad\mbox{and}
     \quad
     |a''_{k}|\leq \frac{|a_{k}|}2\leq
     { \tilde{C}_1}\left\{\begin{array}{cc}
      2^5\varepsilon_1^{-6},
     &  \mbox{if}\  s_0\geq 3,\\
     2^{\frac{6}{v_0}-1} \varepsilon_1^{-\frac{ 6 }{v_0}},
     &  \mbox{if}\ s_0=2.
     \end{array}\right.
$$}
Furthermore,  (\ref{insert 1.1}) together with (\ref{Def hua B})
{  yields the following bound valid  uniformly for   $\bf x\in\mathbb B^d$
\begin{eqnarray*}
   &&
   |h_{3d}({\bf x})-h_{6,d}({\bf x})|
    =
   \left|\sum_{k=1}^3\left[\sum_{\ell=1}^d\frac{a_{k}}2\phi(w_{k}\cdot x^{(\ell)}+\theta_0
       )-\mathcal Bh^*_{2}\left(\frac{\sum_{\ell=1}^da_{k}\phi(w_{k}\cdot x^{(\ell)}+\theta_0
       )}{2\mathcal B}\right)\right]\right|\\
  &\leq&
  3\mathcal
   B\varepsilon_2 =3{ \tilde{C}_1}d\varepsilon_2\left\{\begin{array}{cc}
     2^5 \varepsilon_1^{-6},
     &  \mbox{if}\  s_0\geq 3,\\
     2^{\frac{6}{v_0}-1} \varepsilon_1^{-\frac{ 6 }{v_0}},
     &  \mbox{if}\ s_0=2.
     \end{array}\right.
\end{eqnarray*}}
Setting
$\varepsilon_2=2^{-\frac{6}{v_0}}\frac{1}{3d\tilde{C}_1}\left\{\begin{array}{cc}
      \varepsilon_1^{7},
     &  \mbox{if}\  s_0\geq 3,\\
      \varepsilon_1^{\frac{ 6+v_0 }{v_0}},
     &  \mbox{if}\ s_0=2,
     \end{array}\right. $ the {  above
estimate  yields}
\begin{equation}\label{insertting 1.2}
       |h_{3d}({\bf x})-h_{6,d}({\bf x})|\leq \varepsilon_1,\qquad \forall\  \bf x\in\mathbb B^d,
\end{equation}
and the parameters of $h_{6,d}(\bf x)$ satisfy {
\begin{equation}\label{insertting 1.3}
  0< w_{k}'\leq 1,\qquad |a'_k|,|a''_{k}|\leq  \bar{C}_4\left\{\begin{array}{cc}
      \varepsilon_1^{-48},
     &  \mbox{if}\  s_0\geq 3,\\
      \varepsilon_1^{-\frac{ 6v_0+42 }{v_0}},
     &  \mbox{if}\ s_0=2,
     \end{array}\right.
\end{equation}}
where $\bar{C}_4$ is a constant depending only on $v_0$,
$\tilde{C}_1$ and $d$.
 Based on (\ref{induction 77777}) and (\ref{insertting 1.2}), {  we obtain}
\begin{equation}\label{insertting 1.4}
   ||{\bf x}|^2_2/2-h_{6,d}({\bf x})|\leq
   \frac{d+2}2\varepsilon_1,\qquad\forall\ {\bf x}\in\mathbb B^d.
\end{equation}
Since $\varepsilon_1\leq\frac1{d+2}$, we have
\begin{equation}\label{jiajiajia}
    \|h_{6,d}\|_{L_\infty(\mathbb B^d)}\leq 1.
\end{equation}

{\it Step 3: Constructing the deep net.} Based on (\ref{Def. h6d})
and (\ref{Def H 2 hidden}), we define
{  \begin{eqnarray}\label{Def 3 layered}
       &&H_{3n+3,s+3,6,d,A}:=H_{3n+3,s+3,A}\circ h_{6,d}(\bf x)\\
       &=&
       \sum_{j=1}^{3n+3}a^*_j\phi\left(\sum_{i=1}^{s+3}a^*_{j,i}\phi\left(\sum_{k=1}^{6}a^*_{k,j,i}
       \phi\left(
       \sum_{\ell=1}^d
       a^{**}_{k,j,i}\phi\left(w^*_{k,j,i}  x^{(\ell)}+
       \theta_0\right)+\theta_0\right)
       +\theta^*_{j,i}\right)+\theta_j^*\right). \nonumber
\end{eqnarray}}
In view of (\ref{Jackson for 3}), we get
\begin{eqnarray}\label{p i 3}
      &&|f({\bf x})-H_{3n+3,s+3,6,d,A}({\bf x})|
      \leq
      \tilde{C}_{4}(n\delta_\phi(A)
            +n^{-r}+n\varepsilon)\nonumber\\
      &+&
      |H_{3n+3,s+3,A}(|{\bf x}|^2/2)-H_{3n+3,s+3,A}(h_{6,d}({\bf x}))|,\qquad\forall
      {\bf x}\in\mathbb B^d.
\end{eqnarray}
{
Recalling (\ref{induction 5555}) with $\varepsilon_1=1$ and $|b_{A,t_j}(t)/B_1|\leq 1/4,$ we have
$$
   |h_{s+1,j}(t)|\leq  1,\qquad \mbox{and}\quad \left|\frac{h_{s+1,j}(t)}2+\frac{b_{A_j,t}}{2b_1}\right|\leq 2, \qquad t\in \mathbb J.
$$
This together with  (\ref{Def Hj}) implies
$$
      \left|\sum_{i=1}^{s+3}a^*_{j,i}\phi(w^*_{j,i}t+\theta^*_{j,i})\right|\leq
   2.
$$
}
Thus, from Theorem \ref{Theorem:app phi n s A}, we have, for
$0<t\leq1/2$,
\begin{equation}\label{Bound lip111}
   \left|\sum_{i=1}^{s+3}a^*_{j,i}\phi(w^*_{j,i}t+\theta^*_{j,i})+\theta_j^*\right|\leq
   3+3An+|\theta_0|,\qquad  \quad
   \left|w^*_{j,i}t+\theta^*_{j,i}\right|\leq 5An+|\theta_0|+1.
\end{equation}
Thus, for  $0<\varepsilon<1$, $0<\varepsilon_1<\frac1{d+2}$ and
${\bf x}\in\mathbb B^d$, (\ref{jiajiajia}), (\ref{Bound lip111}),
(\ref{bound weight
  deep1}), (\ref{insertting
1.4}) and $\|\phi'\|_{L^\infty(\mathbb R)}\leq 1$  {  yield}
\begin{eqnarray*}
     &&\left|H_{3n+3,s+3,A}(|{\bf x}|^2/2)-H_{3n+3,s+3,A}(h_{6,d}({\bf x}))\right|\\
     &=&
     \left|\sum_{j=1}^{3n+3}a^*_j\phi\left(\sum_{i=1}^{s+3}a^*_{j,i}\phi(w^*_{j,i} |{\bf x}|^2/2
     +\theta^*_{j,i})+\theta_j^*\right)-\sum_{j=1}^{3n+3}a^*_j\phi\left(\sum_{i=1}^{s+3}a^*_{j,i}\phi(w^*_{j,i}h_{6,d}({\bf x})
     +\theta^*_{j,i})+\theta_j^*\right)
     \right|\\
     & \leq&
      \sum_{j=1}^{3n+3}|a_j^*|\left|\sum_{i=1}^{s+3}a^*_{j,i}\phi(w^*_{j,i}|{\bf x}|^2/2
     +\theta^*_{j,i})-\sum_{i=1}^{s+3}a^*_{j,i}\phi(w^*_{j,i}h_{6,d}({\bf x})
     +\theta^*_{j,i})\right|\\
     &\leq&
      \sum_{j=1}^{3n+3}|a_j^*|\sum_{i=1}^{s+3}|a^*_{j,i}w^*_{j,i}|\left||{\bf x}|^2/2
      - h_{6,d}({\bf x})\right|\\
      &\leq&
      \bar{C}_5 An^2\varepsilon_1\left\{\begin{array}{cc}
      \varepsilon^{-7(s+1)!},
     &  \mbox{if}\quad  s_0\geq 3, s_0> s\\
       \varepsilon^{-\frac{7 }{v_0}(s+1)!},
     &  \mbox{if}\quad s_0=s\geq 3,\\
      \varepsilon^{-\frac{v_0+6}{v_0} (s+1)!},
     &  \mbox{if}\quad  s_0=2, s_0>s,\\
       \varepsilon^{-\frac{v_0+6 }{v_0^2}(s+1)!},
     &  \mbox{if}\quad s_0=s=2,
     \end{array}\right.
\end{eqnarray*}
where $\bar{C_5}\geq 1$ is a constant independent of $\varepsilon$,
$\varepsilon_1$, $n$ or $A$. Now we determine $\varepsilon_1$ by
\begin{equation}\label{varepsilon1}
       \varepsilon_1=\frac{1}{\bar{C}_5(d+2)An^2}\left\{\begin{array}{cc}
      \varepsilon^{1+7(s+1)!},
     &  \mbox{if}\quad  s_0\geq 3, s_0> s\\
       \varepsilon^{1+\frac{7 }{v_0}(s+1)!},
     &  \mbox{if}\quad s_0=s\geq 3,\\
      \varepsilon^{1+\frac{v_0+6}{v_0} (s+1)!},
     &  \mbox{if}\quad  s_0=2, s_0>s,\\
       \varepsilon^{1+\frac{v_0+6 }{v_0^2}(s+1)!},
     &  \mbox{if}\quad s_0=s=2
     \end{array}\right.\leq\frac{1}{(d+2)},
\end{equation}
we have
\begin{equation}\label{p i 5}
         |H_{3n+3,s+3,A}(|{\bf x}|^2)-H_{3n+3,s+3,A}(h_{6,d}({\bf x}))|\leq
         \varepsilon,\qquad\forall\ {\bf x}\in\mathbb B^d.
\end{equation}
Inserting   (\ref{p i 5}) into (\ref{p i 3}) and setting
$\varepsilon=n^{-{r-1}}$, we get
$$
      |f({\bf x})-H_{3n+3,s+3,6,d,A}({\bf x})|
      \leq
      \bar{C}_2(n\delta_\phi(A)
            +n^{-r}),\qquad\forall\
        {\bf x}\in\mathbb B^d,
$$
where $\bar{C}_2$ is a constant independent of $n$ or $A$.

{\it Step 4: Bounding parameters.} Theorem \ref{Theorem:app phi n s
A} with $\varepsilon=n^{-r-1}$ shows that $|\theta_j^*|,
      |\theta^*_{j,i}|\leq  1+3An+|\theta_0|$, and
$$
      |a_j^*|,|a^*_{j,i}|\leq \tilde{C}_{4}\left\{\begin{array}{cc}
      n^{7(r+1)(s+1)!},
     &  \mbox{if}\quad  s_0\geq 3, s_0> s\\
       n^{ \frac{7(r+1) }{v_0}(s+1)!},
     &  \mbox{if}\quad s_0=s\geq 3,\\
      n^{ \frac{(v_0+6)(r+1)}{v_0} (s+1)!},
     &  \mbox{if}\quad  s_0=2, s_0>s,\\
       n^{ \frac{(v_0+6)(r+1) }{v_0^2}(s+1)!},
     &  \mbox{if}\quad s_0=s=2.
     \end{array}\right.
$$
Furthermore, (\ref{Def 3 layered}), (\ref{insertting 1.3}),
(\ref{Def. h6d}), (\ref{varepsilon1}) and $\varepsilon=n^{-r-1}$
shows that {  $|w^*_{k,j,i}|\leq 1$ and
$|a^*_{k,j,i}|,|a^{**}_{k,j,i}|$} can be bounded by
$$
  \bar{C}_1\left\{\begin{array}{cc}
     (An^2)^{48} n^{48r(r+1)(1+7(s+1)!)},
     &  \mbox{if}\quad  s_0\geq 3, s_0> s\\
       (An^2)^{48} n^{48(r+1)(1+\frac{7 }{v_0})(s+1)!},
     &  \mbox{if}\quad s_0=s\geq 3,\\
      (An^2)^{\frac{ 6v_0+42 }{v_0}} n^{\frac{ (6v_0+42)(r+1) }{v_0}(1+\frac{v_0+6}{v_0} (s+1)!)},
     &  \mbox{if}\quad  s_0=2, s_0>s,\\
      (An^2)^{\frac{ 6v_0+42 }{v_0}} n^{\frac{ (6v_0+42)(r+1) }{v_0}(1+\frac{v_0+6}{v_0^2} (s+1)!)},
     &  \mbox{if}\quad s_0=s=2,
     \end{array}\right.
$$
where $\bar{C_1}$ is a constant independent of $A$ or $n$.  This
completes the proof of Theorem \ref{Theorem:upper bound
approximation} {  for $\theta^*_{k,j,i},\theta^*_{k,\ell,j,i}=\theta_0$, $w_{k,\ell,j,i}=w^*_{k,j,i}$ and $a^*_{k,\ell,j,i}=a^{**}_{k,j,i}$.}
$\Box$

To prove Theorem \ref{Theorem:almost optimal} we may apply { Theorem \ref{Theorem:upper bound approximation}}, as follows.

{\bf Proof of Theorem \ref{Theorem:almost optimal}.}
The lower bound is obvious in view of Theorem \ref{Theorem:lower
bound for deep nets 1}. To prove the upper bound, we observe that
under Assumption \ref{Assumption:smooth}, { a  constant
$\bar{C}_6$ depending only on $\phi$ exists  such that
$$
     \delta_\phi(A)\leq \bar{C}_6 A^{-1},\qquad\forall\ A\geq 1.
$$}
Set  $A=n^{r+1}$. Then Assumption \ref{Assumption:smooth} implies {  Assumption \ref{Assumption:smooth1}
with $s_0\geq \max\{3,s+1\}$.} Hence, it follows from Theorem
\ref{Theorem:upper bound approximation}  that {  there exists a deep
net $H_{3n+3,s+3,6,d,A}$ with $|w^*_{k,\ell,j,i}|\leq 1$,
$|\theta^*_{k,\ell,j,i}|,|\theta^*_{k,j,i}|,
|\theta^*_{j,i}|,|\theta_j^*|\leq 1+3n^{r+2}+|\theta_0|$, and
$$
  |a_j^*|,|a^*_{j,i}|,|a^*_{k,j,i}|,|a^*_{k,\ell,j,i}|
  \leq \bar{C}_1
         n^{48 (3+r(r+1)+ r(s+1)!7(r+1))},
$$
such that
$$
       \|f-H_{3n+3,s+3,6,d,A}\|_{L_\infty(\mathbb B^d)}
       \leq \bar{C}_2(n^{-r}+n^{-r}).
$$}
 This completes the proof of Theorem \ref{Theorem:almost optimal} with  $C_3^*=2\bar{C}_2$, $\mathcal R=\max\{|\theta_0|+4,\bar{C}_1\}$ and { $\alpha=48 (3+r(r+1)+ r(s+1)!7(r+1))$}.
$\Box$

\subsection{Proof of Theorem \ref{Theorem: ERM}}

To prove Theorem \ref{Theorem: ERM}, we need the following
well-known oracle inequality that was proved in \cite{Chui2019}.

\begin{lemma}\label{Lemma:oracle}
Let $\rho_X$ be the marginal distribution of  $\rho$ on $\mathcal X$
and $(L^2_{\rho_{_X}}, \|\cdot\|_\rho)$ denote the Hilbert space of
square-integrable functions on $\mathcal X$ with respect to
$\rho_X$.  Set $\mathcal
E_{D}(f):=\frac1m\sum_{i=1}^m(f(x_i)-y_i)^2$, let $\mathcal H$ be a
collection of {  continuous functions   on} $\mathcal X$ and define
\begin{equation}\label{ERM!!!!!!}
        f_{D,\mathcal H}=\arg\min_{f\in\mathcal H}
        \mathcal E_{D}(f).
\end{equation}
Suppose {  there exist constants} $n', \mathcal U>0$, such that
\begin{equation}\label{oracle condition}
    \log \mathcal N(\varepsilon,\mathcal H)\leq
     n'\log \frac{\mathcal U}{\varepsilon},\qquad\forall  \varepsilon>0.
\end{equation}
Then for any ${  h^*}\in\mathcal H$ and {  $\varepsilon>0$},
\begin{eqnarray*}
      &&  Prob\{\|\pi_Mf_{ D,\mathcal H}-f_\rho\|_\rho^2>\varepsilon+2\|{  h^*}-f_\rho\|_\rho^2\}
      \leq
      \exp\left\{n'\log\frac{16\mathcal UM}{\varepsilon}-\frac{3m\varepsilon}{512M^2}\right\}\\
      &+&
      \exp\left\{\frac{-3m\varepsilon^2}{16(3M+\|{  h^*}\|_{L_\infty(\mathcal X)})^2\left(
    6\|{  h^*}-f_\rho\|_\rho^2+
    \varepsilon\right)}\right\}.
\end{eqnarray*}
\end{lemma}

Now we apply Lemma \ref{Lemma:oracle}, Lemma \ref{Lemma:covering
number}, and Theorem \ref{Theorem:almost optimal} {  to prove} Theorem \ref{Theorem: ERM}.

{\bf Proof of Theorem \ref{Theorem: ERM}.}
Let $\mathcal H=\mathcal H_{3,\alpha,\mathcal R}^{tree}$ be the
class of deep nets {  given} in Theorem \ref{Theorem:almost optimal}.
Then, there are totally $\mathcal A_3=\bar{C}_7n$ free parameters in
$\mathcal H=\mathcal H_{3,\alpha,\mathcal R}^{tree}$. Since $|y|\leq
M$ almost surely, we have $\|f_\rho\|_{L_\infty(\mathbb B^d)}\leq
M$. Then, for $f_\rho\in Lip^{(\diamond,r,c_0)}$, {  it follows from  Theorem
\ref{Theorem:almost optimal} that there exists a  $h\in
\mathcal H_{3,\alpha,\mathcal R}^{tree}$  such that
$$
         \|f_\rho-h\|_{L_\infty(\mathbb B^d)}\leq
         \bar{C}_8n^{-r}
$$
where $\bar{C}_7$, $\bar{C}_8$ are constants independent of $n$ and
$\varepsilon$.  It follows that
$$
 \|h\|_{L_\infty(\mathbb B^d)}\leq M+\bar{C}_8.
$$}
By considering $n'=2(\log_2e)\bar{C}_7n$, $\mathcal
U=2^\frac{13}2\mathcal R^{5}(\bar{C}_7n)^{5\alpha}$, {  we see} from
(\ref{covering number estimate:dn}) with $L=3$, $\mathcal
A_L=\bar{C}_7n$ and $c_1=1$ { in Lemma \ref{Lemma:covering number}} that
$$
       \log \mathcal N(\varepsilon,\mathcal H_{3,\alpha,\mathcal R}^{tree})
       \leq n'\log\frac{\mathcal U}{\varepsilon}.
$$
Next {  take} $\bar{C}_9:=\max\{6\bar{C}_8^2,2^{21/2}M\mathcal
R^{5}(\bar{C}_7)^{5\alpha}\}$ and
$\bar{C}_{10}:=\left(\frac{3\bar{C}_9}{2048M^2\bar{C}_7(5\alpha+2r)\log_2e}\right)^{\frac1{2r+1}}.$
{
Note
$$
    2\|h-f_\rho\|_\rho^2\leq 2\|h-f_\rho\|^2_{L_\infty(\mathbb B^d)}\leq 2\bar{C}_8^2n^{-2r}\leq \bar{C}_9n^{-2r}\log n.
$$
Then by setting $n=\left[\bar{C}_{10}m^{\frac1{2r+1}}\right]$,  it
follows from Lemma \ref{Lemma:oracle} with $h^*=h$ that for
\begin{equation}\label{condition for varepsiln}
    \varepsilon\geq \bar{C}_9n^{-2r}\log n\geq 2\|h-f_\rho\|_\rho^2,
\end{equation}}
we have
\begin{eqnarray}\label{probability estimate}
      &&  Prob\{\|\pi_Mf_{
     D,n,\phi}-f_\rho\|_\rho^2>2\varepsilon\}
     \leq
       Prob\{\|\pi_Mf_{
     D,n,\phi}-f_\rho\|_\rho^2>\varepsilon+2\|h-f_\rho\|_\rho^2\}
     \nonumber\\
       &\leq&
      \exp\left\{2(\log_2e)\bar{C}_7n\log\frac{M2^\frac{21}2\mathcal
      R^{5}(\bar{C_7}n)^{5\alpha}}{\varepsilon}-\frac{3m\varepsilon}{512M^2}\right\}\nonumber\\
       &+&
      \exp\left\{\frac{-3m\varepsilon^2}{16(4M+\bar{C}_8)^2\left(
      6
    \bar{C}_8^2n^{-2r}+
    \varepsilon\right)}\right\}\nonumber\\
    &\leq&
    \exp\left\{2(\log_2e)\bar{C}_7(5\alpha+2r)n\log n-\frac{3m\varepsilon}{512M^2}\right\}
    +
    \exp\left\{\frac{-3m\varepsilon}{32(4M+\bar{C}_8)^2}\right\}\nonumber\\
    &\leq&
    \exp\left\{-\frac{3m\varepsilon}{1024M^2}\right\}
    +\exp\left\{-\frac{-3m\varepsilon}{32(4M+\bar{C}_8)^2}\right\}
    \leq 2\exp\left\{-\frac{3m\varepsilon}{64(4M+\bar{C}_8)^2}\right\}\nonumber\\
    &\leq&
      3\exp\left\{-\frac{3m^{\frac{2r}{2r+1}}\varepsilon}{2[64(4M+\bar{C}_8)^2+3\bar{C}_{9}(\bar{C}_{10})^{-2r}]\log (n+1)}\right\}.
\end{eqnarray}
 Then setting
$$
        3\exp\left\{-\frac{3m^{\frac{2r}{2r+1}}\varepsilon}{2[64(4M+\bar{C}_8)^2+3\bar{C}_9(\bar{C}_{10})^{-2r}]\log (n+1)}\right\}=\delta,
$$
we obtain
$$
      \varepsilon=\frac23[64(4M+\bar{C}_8)^2+3\bar{C}_9(\bar{C}_{10})^{-2r}]m^{-\frac{2r}{2r+1}}\log(n+1)
      \log\frac3\delta,
$$
which satisfies (\ref{condition for varepsiln}). Thus, it follows
from (\ref{probability estimate}) that with confidence {  at least}
$1-\delta$, we have
$$
       \|\pi_Mf_{
     D,n,\phi}-f_\rho\|_\rho^2
     \leq C_5^*m^{-\frac{2r}{2r+1}}\log(n+1)
      \log\frac3\delta
      \leq
      C_5^*m^{-\frac{2r}{2r+1}}\log (m+1) \log\frac3\delta,
$$
where
$C_5^*:=\frac83[64(4M+\bar{C}_8)^2+3\bar{C}_9(\bar{C}_{10})^{-2r}]$.
This proves (\ref{learning rate})  by noting the well-known relation
\begin{equation}\label{equality}
        \mathcal E(f)-\mathcal E(f_\rho)=\|f-f_\rho\|_\rho^2.
\end{equation}

To prove the upper bound of (\ref{almost optimal learning rate}), we
may apply the formula
\begin{equation}\label{expectationdef}
      E[\xi]=\int_0^\infty{ Prob}[\xi>t]dt
\end{equation}
with $\xi = \mathcal E(\pi_Mf_{D,n,\phi})-\mathcal E(f_\rho)$. From
(\ref{condition for varepsiln}), (\ref{probability estimate}) and
(\ref{expectationdef}), we have
\begin{eqnarray*}
   &&E\left\{\mathcal E(\pi_Mf_{D,n,\phi})-\mathcal E(f_\rho)\right\}
   =
   \int_0^\infty{ Prob}[\mathcal E(\pi_Mf_{D,n,\phi})-\mathcal
   E(f_\rho)>\varepsilon]d\varepsilon\\
   &=&
   \left(\int_0^{\bar{C}_9n^{-2r}\log
   n}+\int_{\bar{C}_6n^{-2r}\log n}^\infty\right)Prob[\mathcal E(\pi_Mf_{D,n,\phi})-\mathcal
   E(f_\rho)>\varepsilon]d\varepsilon\\
   &\leq&
   \bar{C}_9n^{-2r}\log
   n +3\int_{\bar{C}_9n^{-2r}\log n}^\infty \exp\left\{-\frac{3m^{\frac{2r}{2r+1}}
   \varepsilon}{2[64(4M+\bar{C}_8)^2+3\bar{C}_9(\bar{C}_{10})^{-2r}]\log (n+1)}\right\}
   d\varepsilon\\
   &\leq&
   \bar{C}_9n^{-2r}\log
   n +6[64(4M+\bar{C}_8)^2+3\bar{C}_9(\bar{C}_{10})^{-2r}]m^{-\frac{2r}{2r+1}}\log(n+1)\int_0^\infty
   e^{-t}dt\\
   &\leq&
    C_7^*m^{-\frac{2r}{2r+1}}\log (m+1),
\end{eqnarray*}
where
$$
     C_7^*=6[64(4M+\bar{C}_8)^2+3\bar{C}_9(\bar{C}_{10})^{-2r}]+\bar{C}_9[(\bar{C}_{10})^{-2r}+1].
$$
Finally, to prove the lower bound of (\ref{almost optimal learning
rate}), we note that since ${\bf x}_1,\dots, {\bf x}_m$ are
independent random variables, so are $|{\bf x}_1|^2, \dots, |{\bf
x}_m|^2$. Thus, the data set $\{(|{\bf x}_i|^2, y_i)\}_{i=1}^m$ is
independently drawn according to some distribution {  $\rho$} defined
on $\mathbb I\times [-M,M]$.  From \cite[Theorem 3.2]{Gyorfi2002},
there exists some $\rho_0'$ with the regression function $g_\rho\in
Lip^{(r,c_0)}_{\mathbb I}$, such that the learning rates of all
estimates based on $m$ {  sample points} are not smaller than
$C_6^*m^{-\frac{2r}{2r+1}}$.  Setting $f_\rho({\bf x}) =g_\rho(|{\bf
x}|^2)$, we may conclude that the lower bound of (\ref{almost
optimal learning rate}) holds. This completes the proof of Theorem
\ref{Theorem: ERM}.
$\Box$

\section{Proof of Lemma \ref{Lemma:Relation c and l}}\label{Sec. Auxiliary Lemmas}

The proof of Lemma \ref{Lemma:Relation c and l}  depends on the
following two {  lemmas. They involve the
$\varepsilon$-packing number of $V$  defined by}
$$
   \mathcal M(\varepsilon,V):=\max\{s:\exists f_1,\dots, f_s\in V,
   \|f_i-f_j\|_{L_\infty(\mathbb B^d)}\geq\varepsilon,\forall i\neq j\}
$$
 The first {  lemma which can be found in \cite[Lemma 9.2]{Gyorfi2002}) establishes a trivial relation between $\mathcal N(\varepsilon,V)$ and $\mathcal M(\varepsilon,V)$.}

\begin{lemma}\label{Claim:covering and packing}
      For  $\varepsilon>0$ and  $V\subseteq L_\infty(\mathbb B^d)$, {  we have}
$$
     \mathcal M(2\varepsilon,V)\leq \mathcal N(\varepsilon,V)\leq \mathcal
     M(\varepsilon,V).
$$
\end{lemma}

{  To state the second lemma, for} $N^*\in\mathbb N$, consider
the set $E^{N^*}$  of all vectors
$\epsilon:=(\epsilon_1,\dots,\epsilon_{N^*})$ for
$\epsilon_1,\dots,\epsilon_{N^*}=\pm1,$ so that the cardinality
$|E^{N^*}|$ of the set $E^{N^*}$ is given by
\begin{equation}\label{Cardinality}
       |E^{N^*}|=2^{N^*}.
\end{equation}
Let $\tilde{g}$ be a real-valued compactly supported function that
vanishes outside $(-1/2,1/2)$ and satisfies both $\max_{t\in
[-1/2,1/2]}|\tilde{g}(t)|= c_0/2$ and $\tilde{g}\in
Lip^{(r,c_02^{v-1})}_{\mathbb R}.$ Also, partition the unit interval
$\mathbb I$ as the union of $N^*$ pairwise disjoint sub-intervals
$A_j$ of equal length $1/N^*$ and centers at  $\{\xi_j\}$ for
$j=1,\cdots,N^*$. Consider the dilated/scaled functions
$\tilde{g}_j(t):=({N^*})^{-r}\tilde{g}({N^*}(t-\xi_j))$ defined on
$\mathbb I$.  Then based on the set
\begin{equation}\label{subset}
     \mathcal G_E:=\left\{g^*(t)=\sum_{j=1}^{N^*}\epsilon_j\tilde{g}_j(t):\epsilon=(\epsilon_1,\dots,\epsilon_{N^*})\in
     E^{N^*}\right\}
\end{equation}
of univariate functions, we introduce the collection
\begin{equation}\label{functional class}
    \mathcal F_E:=\left\{f({\bf x})=g(|{\bf x}|^2): g\in\mathcal G_E\right\}.
\end{equation}
of radial functions defined on the $\mathbb B^d$. {
\begin{lemma}\label{Claim:distance}
Let ${N^*}\in\mathbb N$. Then
\begin{equation}\label{subseteq111}
      \mathcal F_E\subset Lip^{(\diamond,r,c_0)}.
\end{equation}
and in addition, for any $f\neq f_1\in\mathcal F_E$,
\begin{equation}\label{distance111}
           \|f-f_1\|_{L_\infty(\mathbb B^d)}\geq c_0({N^*})^{-r}.
\end{equation}
\end{lemma}}

{\bf Proof.}
To prove {  (\ref{subseteq111})}, observe that since
$$
          |{N^*}(t-\xi_j)-{N^*}(t-\xi_{j'})|
          ={N^*}|\xi_j-\xi_{j'}|\geq1,\qquad \forall\ j\neq j',
$$
it is not possible for both ${N^*}(t-\xi_j)$ and ${N^*}(t-\xi_{j'})$
to be in $(-1/2,1/2)$. Hence, it follows from the support assumption
$supp(\tilde{g})\subset (-1/2,1/2)$ of $\tilde{g}$ that for {  an}
arbitrary $t\in\mathbb I$, there is at most one
$j_0\in\{1,2,\dots,N^*\}$ such that $(\tilde{g}_{j_0})^{(s)}(t)\neq
0$. Then the justification of (\ref{subseteq111}) can be argued in
two separate cases.

First, for {  an} arbitrary $g^*\in \mathcal G_E$, if $t,t'\in A_{j_1}$ for
some $j_1\in\{1,2,\dots,{N^*}\}$, then in view of
$supp({\tilde{g}})\subset (-1/2,1/2)$, $r=s+v$, $|\epsilon_j|=1$,
and $\tilde{g}\in Lip^{(r,c_02^{-1+v})}_{\mathbb R}$, we have
$$
  (\tilde{g})^{(s)}({N^*}(t-\xi_{j}))=(\tilde{g})^{(s)}({N^*}(t'-\xi_{j}))=0,\qquad\forall\ j\neq j_1
$$
and
\begin{eqnarray*}
      &&|(g^*)^{(s)}(t)-(g^*)^{(s)}(t')|
      =\left|\sum_{j=1}^{N^*}\epsilon_j[(\tilde{g}_j)^{(s)}(t)-(\tilde{g}_{j})^{(s)}(t')]\right|\\
      &\leq&
      ({N^*})^{-r+s}\left|\sum_{j=1}^{N^*}\epsilon_j[(\tilde{g})^{(s)}({N^*}(t-\xi_j))-(\tilde{g})^{(s)}({N^*}(t'-\xi_j))]\right|\\
      &=&
       ({N^*})^{-r+s}\left|\epsilon_{j_1}[(\tilde{g})^{(s)}({N^*}(t-\xi_{j_1}))-(\tilde{g})^{(s)}({N^*}(t'-\xi_{j_1}))]\right|\\
      &\leq&
      ({N^*})^{-r+s}c_02^{-1+v}|{N^*}(t-\xi_{j_1})-{N^*}(t'-\xi_{j_1})|^v
      \leq
      c_0|t-t'|^v.
\end{eqnarray*}

Next, if $t\in A_{j_2}$ and $t'\in A_{j_3}$ {  with} $j_2\neq j_3$,
then
$$
  (\tilde{g})^{(s)}({N^*}(t-\xi_{j}))=(\tilde{g})^{(s)}({N^*}(t-\xi_{j'}))=0,\qquad\forall
  j\neq j_2, j'\neq j_3.
$$
We may choose the endpoints  $\eta_{j_2}\in A_{j_2}$ and
$\eta_{j_3}\in A_{j_3}$ so that $\eta_{j_2}$ and $\eta_{j_3}$ are on
the segment between $t$ and $t'$. This together with
$supp(\tilde{g})\subset[-1,2,1/2]$ implies that
$$
          |t-\eta_{j_2}|+|t'-\eta_{j_3}|\leq |t-t'|
$$
and
$$
       (\tilde{g})^{(s)}({N^*}(\eta_{j_2}-\xi_{j_2}))=(\tilde{g})^{(s)}({N^*}(\eta_{j_3}-\xi_{j_3}))=0.
$$
Thus, it follows from $r=s+v$, $|\epsilon_j|=1$, $\tilde{g}\in
Lip^{(r,c_02^{-1+v})}$ and Jensen's inequality, that
\begin{eqnarray*}
         &&|(g^*)^{(s)}(t)-(g^*)^{(s)}(t')|
      =\left|\sum_{j=1}^{N^*}\epsilon_j[(\tilde{g}_j)^{(s)}(t)-(\tilde{g}_{j})^{(s)}(t')]\right|\\
      &=&
      ({N^*})^{-r+s}\left|\sum_{j=1}^{N^*}\epsilon_j[(\tilde{g})^{(s)}({N^*}(t-\xi_j))-(\tilde{g})^{(s)}({N^*}(t'-\xi_j))]\right|\\
        &\leq&
      ({N^*})^{-r+s}\left|(\tilde{g})^{(s)}({N^*}(t-\xi_{j_2}))\right|+({N^*})^{-r+s}\left|(\tilde{g})^{(s)}({N^*}(t'-\xi_{j_3}))\right|\\
      &=&
      ({N^*})^{-r+s} \left|(\tilde{g})^{(s)}({N^*}(t-\xi_{j_2}))-(\tilde{g})^{(s)}({N^*}(\eta_{j_2}-\xi_{j_2}))\right|\\
      &+&
      ({N^*})^{-r+s}\left|(\tilde{g})^{(s)}({N^*}(t'-\xi_{j_3}))-(\tilde{g})^{(s)}({N^*}(\eta_{j_3}-\xi_{j_3}))\right|\\
      &\leq&
       c_02^{v-1}\left[|t-\eta_{j_2}|^v+|t'-\eta_{j_3}|^v\right]
       =
       c_02^{v}\left[\frac{|t-\eta_{j_2}|^v+|t'-\eta_{j_3}|^v}2\right]\\
      &\leq&
      c_02^{v}\left[\frac{|t-\eta_{j_2}|+|t'-\eta_{j_3}|}2\right]^v
      \leq
      c_02^{v}\left[\frac{|t-t'|}2\right]^v
      = c_0|t-t'|^v.
\end{eqnarray*}
{ From the above arguments, we know that  (\ref{subseteq111}) holds in view
of (\ref{functional class}).}

Finally, to prove (\ref{distance111}), let $f,f_1\in \mathcal F_E$
be two different functions. Then there exist
$\epsilon,\epsilon'\in E^{N^*}$ with $\epsilon\neq\epsilon'$ such
that
$$
    f({\bf x})-f_1({\bf x})=\sum_{j=1}^{N^*}\epsilon_j\tilde{g}_j(|{\bf x}|^2)-\sum_{j=1}^{N^*}\epsilon'_j\tilde{g}_j(|{\bf x}|^2)
    =({N^*})^{-r}\sum_{j=1}^{N^*}(\epsilon_j-\epsilon'_j)\tilde{g}({N^*}(|{\bf x}|^2-\xi_j)).
$$
Therefore, we have
\begin{eqnarray*}
      &&\|f-f_1\|_{L_\infty(\mathbb B^d)}
       =
      ({N^*})^{-r}\max_{{\bf x}\in\mathbb
      B^d}\left|\sum_{j=1}^{N^*}(\epsilon_j-\epsilon'_j)\tilde{g}({N^*}(|{\bf x}|^2-\xi_j))\right|\\
      &=&
      ({N^*})^{-r}\max_{t\in\mathbb
      I}\left|\sum_{j=1}^{N^*}(\epsilon_j-\epsilon'_j)\tilde{g}({N^*}(t-\xi_j))\right|
      \\
       &=&
      ({N^*})^{-r}\max_{j'=1,2\dots,{N^*}}\max_{t\in A_{j'}}\left|\sum_{j=1}^{N^*}(\epsilon_j-\epsilon'_j)\tilde{g}({N^*}(t-\xi_j))\right|\\
      &=&
      ({N^*})^{-r}\max_{j'=1,2\dots,{N^*}}\max_{t\in A_{j'}}\left|(\epsilon_{j'}-\epsilon'_{j'})\tilde{g}({N^*}(t-\xi_{j'}))\right|\\
      &=&
      ({N^*})^{-r}\max\left\{\max_{j':\epsilon_{j'}-\epsilon'_{j'}=2}\max_{t\in
      A_{j'}}\left|2\tilde{g}({N^*}(t-\xi_{j'}))\right|,\max_{j':\epsilon_{j'}-\epsilon'_{j'}=-2}\max_{t\in
      A_{j'}}\left|-2\tilde{g}({N^*}(t-\xi_{j'}))\right|\right\}.
\end{eqnarray*}
Noting that $\{t={N^*}(\tau-\xi_j):\tau\in A_j\}=[-1/2,1/2]$ for
each $j\in\{1,\dots,N^*\}$ and $
\max_{t\in[-1/2,1/2]}|\tilde{g}(t)|=c_0/2$, we obtain
$$
      \|f-f_1\|_{L_\infty(\mathbb B^d)}
     =
      2(N^*)^{-r}\max_{t\in [-1/2,1/2]}|\tilde{g}(t)|
       =
       c_0({N^*})^{-r}.
$$
Thus, (\ref{distance111}) holds. This completes the proof of {  Lemma}
\ref{Claim:distance}.
$\Box$

We now return to the proof of Lemma \ref{Lemma:Relation c and l}.

{\bf Proof of Lemma \ref{Lemma:Relation c and l}.}
Let $\nu>0$ to be determined later, and denote
\begin{equation}\label{def delta}
     \delta:=\delta_\nu:=\mbox{dist}(\mathcal
    F_E,V,L_\infty(\mathbb B^d))+\nu.
\end{equation}
For every $f\in\mathcal F_E,$ choose a function $Pf\in V$,  so that
\begin{equation}\label{projection}
        \|f-Pf\|_{L_\infty(\mathbb
        B^d)}\leq\delta.
\end{equation}
Observe that $Pf$ is {  not necessarily} unique. Define $\mathcal
S_E:=\{Pf:f\in\mathcal F_E\}\subseteq V$. Then for $f^*=Pf$ and
$f^*_1=Pf_1$ with  $f\neq f_1\in \mathcal F_E$, we have
\begin{eqnarray*}
     \|f^*-f_1^*\|_{L_\infty(\mathbb B^d)}
     &=&
     \|Pf-Pf_1\|_{L_\infty(\mathbb B^d)}
     =
     \|Pf-f+f-f_1+f_1-Pf_1\|_{L_\infty(\mathbb B^d)}\\
     &\geq&
     \|f-f_1\|_{L_\infty(\mathbb B^d)}
     -
     \|Pf-f\|_{L_\infty(\mathbb B^d)}
     -\|Pf_1-f_1\|_{L_\infty(\mathbb B^d)},
\end{eqnarray*}
which together with  (\ref{distance111}) {  implies}
\begin{equation}\label{1.lower1}
   \|f^*-f_1^*\|_{L_\infty(\mathbb B^d)}\geq c_0(N^*)^{-r}-2\delta.
\end{equation}
We claim that $\delta>\frac{c_0}4(N^*)^{-r}$, where $N^*$ is given
by
\begin{equation}\label{Def:Nstar}
    N^*=\left[(\beta+2r+4)N\log_2(2C_1'+4C_2'(\beta+2r+4)^r/c_0+N)\right].
\end{equation}
To prove the claim, suppose to the contrary that
\begin{equation}\label{condration condition}
     \delta\leq \frac{c_0}4(N^*)^{-r}.
\end{equation}
 Then (\ref{1.lower1}) implies
$$
        \|f^*-f^*_1\|_{L_\infty(\mathbb B^d)}\geq \frac{c_0}2(N^*)^{-r}.
$$
That is, $Pf\neq Pf_1$ is consequence of $f\neq f_1$, so that in
view of (\ref{Cardinality}),
$$
     |\mathcal S_E|=|\mathcal F_E|=  |E^{N^*}| =2^{N^*}.
$$
Consider $\varepsilon_0=\frac{c_0}2(N^*)^{-r}$. Then we obtain
$$
    \mathcal
     M(\varepsilon_0,V)\geq 2^{N^*}.
$$
On the other hand, since $ \mathcal S_E\subseteq V$, it follows from
(\ref{covering condition}) and {  Lemma} \ref{Claim:covering and
packing} that
$$
      \mathcal
     M(\varepsilon_0,V)\leq \mathcal
     N(\varepsilon_0/2,V)\leq
     C_1'\left(\frac{2C_2'N^\beta}{\varepsilon_0}\right)^N
     =C_1'\left(4C_2'N^\beta (N^*)^r/c_0\right)^N.
$$
Combining the above two inequalities, we have
\begin{equation}\label{number packing}
        2^{N^*}\leq C_1'\left({4C_2'N^\beta
        (N^*)^r}/c_0\right)^N.
\end{equation}
The choice of $N^*$ in (\ref{Def:Nstar}) tells us that (\ref{number
packing}) holds, but it
 implies that
\begin{eqnarray*}
    &&(\beta+2r+4)N\log_2(2C_1'+4\tilde{C}_2(\beta+2r+4)^r/c_0+N)
     \leq
     N\log_2(4C_2'(\beta+2r+4)^r/c_0)\\
     &+&
     \log_2 (2C_1')+N(\beta+r)\log_2
    N+rN\log_2\log_2((2C_1'+4C_2'(\beta+2r+4)^r/c_0+N))\\
    &\leq&
    (\beta+2r+3)N\log_2(2C_1'+4C_2'(\beta+2r+4)^r/c_0+N),
\end{eqnarray*}
which is a { contradiction. This verifies   our claim, so}
$$
       \delta>\frac{c_0(N^*)^{-r}}4=
       \frac{c_0}4\left[(\beta+2r+4)N\log_2(2C_1'+4C_2'(\beta+2r+4)^r/c_0+N)\right]^{-r}.
$$
{  Now, we   determine $\nu$ by $\nu=\mbox{dist}(\mathcal
F_E,V,L_\infty(\mathbb B^d))$. Then $\nu=\frac\delta2$ by (\ref{def
delta}), and we obtain}
$$
   \mbox{dist}(\mathcal
    F_E,V,L_\infty(\mathbb B^d))=\frac\delta2>\frac{c_0}8\left[(\beta+2r+4)N\log_2(2C_1'+4C_2'(\beta+2r+4)^r/c_0+N)\right]^{-r}.
$$
In view of (\ref{subseteq111}), we have
$$
   \mbox{dist}(Lip^{(\diamond,r,c_0)},V,L_\infty(\mathbb B^d))\geq \mbox{dist}(\mathcal
    F_E,V,L_\infty(\mathbb B^d))\geq
   C_3'\left[N\log_2(N+C_4')\right]^{-r}
$$
with $C_3'=\frac{c_0}8(\beta+2r+4)^{-r}$ and
$C_4'=2C_1'+4C_2'c_0^{-1}(\beta+2r+4)^r$. This completes the proof
of Lemma \ref{Lemma:Relation c and l}.
$\Box$


\begin{thebibliography}{99}


\bibitem{Buhmann2003} M. D. Buhmann, Radial Basis Functions: Theory and
Implementation, Cambridge Monograph on Appl. and Comput. Math. Vol
12, Cambridge Univ. Press, 2003.

\bibitem{Chen1993}
D. B. Chen, Degree of approximation by superpsitions of a sigmoidal
function, Approx. Theory \& its Appl., 9 (1993), 17-28.

\bibitem{Chui1994}
C. K. Chui, X. Li, H. N. Mhaskar, Neural networks for lozalized
approximation, Math. Comput., 63 (1994), 607-623..

\bibitem{Chui2016}
  C. K. Chui, H. N. Mhaskar, Deep nets for local manifold
learning, Front. Appl. Math. Stat.,  4 (2018): 14.

\bibitem{Chui2017}
C. K. Chui, S. B. Lin, D. X. Zhou, Construction of neural networks
for realization of localized deep
  learning, Froint. Appl. Math. { Stat.}, 4 (2018): 12.

\bibitem{Chui2019}
C. K. Chui, S. B. Lin, D. X. Zhou, Generalization capability of deep nets with tree structures, Froint. Appl. Math. {  Stat.}, Submitted.

\bibitem{Cucker2007}
F. Cucker, D. X. Zhou. Learning Theory: An Approximation Theory
Viewpoint,  Cambridge University Press, Cambridge, 2007.

%

\bibitem{Goodfellow}
I. Goodfellow, Y. Bengio, A. Courville, {Deep Learning}, MIT Press,
2016.

{
\bibitem{Guo2019}
Z. C. Guo, L. Shi, and S. B. Lin. Realizing data features by deep nets, arXiv repreprint arXiv: 1901.00130, 2019.
}

\bibitem{Gyorfi2002}
L. Gy\"{o}rfy, M. Kohler, A. Krzyzak, H. Walk, A Distribution-Free
Theory of Nonparametric Regression, Springer, Berlin, 2002.

\bibitem{Hinton2006}
G. E. Hinton, S. Osindero, Y. W. Teh, A fast learning algorithm for
deep belief netws, Neural Comput., 18 (2006), 1527-1554.

\bibitem{Kohler2017}
M. Kohler, A. Krzyzak, Nonparametric regression based on
hierarchical interaction models, IEEE Trans. Inform. Theory, 63 (2017),
1620-1630.

\bibitem{Konovalov2008}
V. N. Konovalov, D. Leviatan, V. E. Maiorov, Approximation by
polynomials and ridge functions of classes of s-monotone radial
functions, J. Approx. Theory,  152 (2008), 20-51.

\bibitem{Konovalov2009}
V. N. Konovalov, D.  Leviatan, V. E. Maiorov, Approximation of
Sobolev classes by polynomials and ridge functions, J. Approx.
Theory,   159 (2009), 97-108.

\bibitem{Lin2016}
S. B. Lin, X. Guo, and D. X. Zhou. Distributed learning with
regularized least
  squares. J. Mach. Learn. Res.,   18 (2017) (92), 1-31.
%

\bibitem{Lin2018}
S. B. Lin, D. X. Zhou, Distributed kernel-based gradient descent
algorithms, Constr. Approx., 47 (2018), 249-276.

\bibitem{Lin2018a}
S. B. Lin, Generalization and expressivity for deep nets, IEEE
Trans. Neural Netw. Learn. Syst., To Appear.

\bibitem{Linh2017}
H. W. Lin, M. Tegmark, D. Rolnick, Why does deep and cheap learning
work so well?, J. Stat. Phys., 168 (2017), 1223-1247.

\bibitem{Maiorov1999b}
V. Maiorov, A. Pinkus, Lower bounds for approximation by MLP neural
networks, Neurocomputing,  25 (1999), 81-91.

\bibitem{Maiorov1999c}
V. Maiorov, J. Ratsaby, On the degree of approximation by manifolds
of finite pseudo-dimension, Constr. Approx., 15 (1999), 291-300.

\bibitem{McCane2017}
B. McCane, L. Szymanski, Deep radial kernel networks: approximating
radially symmetric functions with deep networks, arXiv preprint
arXiv:1703.03470, 2017.

\bibitem{Meylan2006}
L. Meylan, S. Susstrunk, High dynamic range image rendering with a
retinex-based adaptive filter, IEEE Trans. { Image Proc.},   15 (2006),
2820-2830.

\bibitem{Mhaskar1993}
H. N. Mhaskar, Approximation properties of a multilayered
feedforward artificial neural network, Adv. Comput. Math., 1 (1993),
61-80.

\bibitem{Mhaskar1996}
H. N. Mhaskar, Neural networks for optimal approximation of smooth
and analytic functions, Neural Comput., 8 (1996), 164-177.

\bibitem{Mhaskar2004} H. N. Mhaskar, When is approximation by Gaussian
networks necessarily a linear process?, Neural Networks, 17 (2004),
989 -1001.

\bibitem{Mhaskar2016a}
H. N. Mhaskar, T. Poggio, Deep vs. shallow networks: An
approximation theory perspective, Anal. Appl., 2016 (14), 829-848.

\bibitem{Qi2017}
C. R. Qi, H. Su, K. Mo, L. J. Guibas,   PointNet: deep learning on
point sets for 3D classification and segmentation, {  CVPR, pp.
77-85, 2017}.

\bibitem{Satriano2011}
C. Satriano, Y. M. Wu, A. Zollo, H. Kanamori, Earthquake early
warning: Concepts, methods and physical grounds, Soil Dynamics
Earth. Engineer., 31 (2011), 106-118.

\bibitem{Shaham2015}
U. Shaham,  A. Cloninger, R. R. Coifman, Provable approximation
properties for deep neural networks, Appl. Comput. Harmon. Anal., To
appear.

%

\bibitem{Wu2005}
Q. Wu, D. X. Zhou, SVM soft margin classifiers: linear programming
versus quadratic programming,  Neural Comput.,  17 (2015),
1160-1187.

\bibitem{Wu2006}
Q. Wu, Y. Ying, D. X. Zhou, Learning rates of least-square
regularized regression, Found.  Comput. Math.,   6 (2006), 171-192.

\bibitem{Xie2010}
T. Xie, F. Cao, The errors in simultaneous approximation by
feed-forward neural networks, Neurocomputing,  73(2010), 903-907.

\bibitem{Yarotsky2017}
D. Yarotsky, Error bounds for aproximations with deep ReLU networks,
Neural Networks, 94 (2017), 103-114.

\bibitem{Ying2017}
Y. Ying, D. X. Zhou, Unregularized online learning algorithms with general loss functions. Appl. Comput. {  Harmonic}  Anal.,   42 (2017), 224-244.


\bibitem{Ye2008}
G. B. Ye, D. X. Zhou, Learning and approximation by Gaussians on
Riemannian manifolds, Adv. Comput. Math., 29 (2008), 291-310.

\bibitem{Zhou2002}
D. X. Zhou, The covering number in learning theory, J. {  Complexity},
  18 (2002), 739-767.

\bibitem{Zhou2003}
 D. X. Zhou, Capacity of reproducing kernel spaces in learning
 theory,
IEEE Trans. { Inform.} Theory,  49 (2003), 1743-1752.

\bibitem{Zhou2006}
{D. X. Zhou, K. Jetter, Approximation with polynomial kernels and
SVM classifiers, Adv. Comput. Math., 25 (2006), 323-344.}


\bibitem{Zhou2018}
D. X. Zhou, Deep distributed convolutional neural networks:
Universality, Anal. Appl., 16 (2018), 895-919.



\bibitem{Zhou2018a}
D. X. Zhou. Universality of deep donvolutional neural networks,
 arXiv:1805.10769, 2018.

\end{thebibliography}
\end{document}